\documentclass{article}

\PassOptionsToPackage{numbers}{natbib}
\usepackage[preprint]{neurips_2023}

\usepackage[utf8]{inputenc} % allow utf-8 input
\usepackage[T1]{fontenc}    % use 8-bit T1 fonts
\usepackage{hyperref}       % hyperlinks
\usepackage{url}            % simple URL typesetting
\usepackage{booktabs}       % professional-quality tables
\usepackage{amsfonts}       % blackboard math symbols
\usepackage{nicefrac}       % compact symbols for 1/2, etc.
\usepackage{microtype}      % microtypography
\usepackage{xcolor}         % colors
\usepackage{adjustbox}
\usepackage{multirow}
\usepackage{subcaption}
\usepackage{colortbl}
\usepackage{xspace}
\usepackage{lipsum}
\usepackage{enumitem}
\usepackage{amsmath}
\usepackage{duckuments}
\usepackage{algpseudocode}  
\usepackage{setspace}       
\usepackage{scrextend}
\usepackage{mathtools}
\usepackage{graphicx}
\usepackage{subcaption}
\usepackage{enumitem}
\usepackage{colortbl}
\usepackage{wrapfig}

\usepackage{soul}
\usepackage{xcolor}
\usepackage{listings}

\usepackage{etoolbox}
\usepackage{algorithm}
\makeatletter
\AfterEndEnvironment{algorithm}{\let\@algcomment\relax}
\AtEndEnvironment{algorithm}{\kern2pt\hrule\relax\vskip3pt\@algcomment}
\let\@algcomment\relax
\newcommand\algcomment[1]{\def\@algcomment{\footnotesize#1}}
\renewcommand\fs@ruled{\def\@fs@cfont{\bfseries}\let\@fs@capt\floatc@ruled
  \def\@fs@pre{\hrule height.8pt depth0pt \kern2pt}%
  \def\@fs@post{}%
  \def\@fs@mid{\kern2pt\hrule\kern2pt}%
  \let\@fs@iftopcapt\iftrue}
\makeatother
\definecolor{pinegreen}{rgb}{0.0, 0.47, 0.44}
\definecolor{cornellred}{rgb}{0.7, 0.11, 0.11}
\definecolor{cadmiumgreen}{rgb}{0.0, 0.42, 0.24}
\definecolor{royalblue}{rgb}{0.0, 0.14, 0.4}
\definecolor{spirodiscoball}{rgb}{0.06, 0.75, 0.99}
\definecolor{mylightblue}{rgb}{0.85, 0.90, 0.94}
\definecolor{kaistblue}{RGB}{20,135,200}
\definecolor{auburn}{RGB}{166,38,57}
\definecolor{Gray}{gray}{0.9}
\definecolor{BrickRed}{rgb}{0.6,0,0}
\definecolor{RoyalBlue}{rgb}{0,0,0.8}
\definecolor{Tdgreen}{rgb}{0,0.4,0.7}
\definecolor{darkblue}{rgb}{0,0.08,0.45}
\hypersetup{citecolor=darkblue, linkcolor=darkblue}

\hypersetup{
  urlcolor = cadmiumgreen,
  linkcolor = cornellred,
  citecolor  = kaistblue,
  colorlinks = true,
}

\title{
Collaborative Score Distillation \\
for Consistent Visual Synthesis
}

\newcommand{\sname}{CSD\xspace} % short name
\newcommand{\ssname}{CSD-Edit\xspace} % short name

\author{%
Subin Kim\thanks{Equal contribution}$^{\ \,1}$
\qquad Kyungmin Lee\footnotemark[1]$^{\ \,1}$
\qquad June Suk Choi$^{1}$
\qquad {Jongheon Jeong}$^{1}$  \vspace{0.3em} \\
\textbf{Kihyuk Sohn}$^{2}$ 
\qquad \textbf{Jinwoo Shin}$^{1}$ \vspace{0.3em} \\
$^1$KAIST \qquad $^2$Google Research \vspace{0.3em} \\
\footnotemark[1]\,\,\texttt{\{subin-kim, kyungmnlee\}@kaist.ac.kr}
}

\begin{document}

\maketitle
 
\begin{abstract}
Generative priors of large-scale text-to-image diffusion models enable a wide range of new generation and editing applications on diverse visual modalities. However, when adapting these priors to complex visual modalities, often represented as multiple images (e.g., video), achieving consistency across a set of images is challenging.
In this paper, we address this challenge with a novel method, Collaborative Score Distillation (\sname).
{\sname} is based on the Stein Variational Gradient Descent (SVGD). Specifically, we propose to consider multiple samples as ``particles'' in the SVGD update and combine their score functions to distill generative priors over a set of images synchronously.
Thus, \sname facilitates seamless integration of information across 2D images, leading to a consistent visual synthesis across multiple samples.
We show the effectiveness of {\sname} in a variety of tasks, encompassing the visual editing of panorama images, videos, and 3D scenes.
Our results underline the competency of \sname as a versatile method for enhancing inter-sample consistency, thereby broadening the applicability of text-to-image diffusion models.\footnote{Visualizations are available at the website \url{https://subin-kim-cv.github.io/CSD}.}
\end{abstract}

\section{Introduction}\label{sec:intro}
\begin{figure*}[t]
\centering\small
\includegraphics[width=1\textwidth]{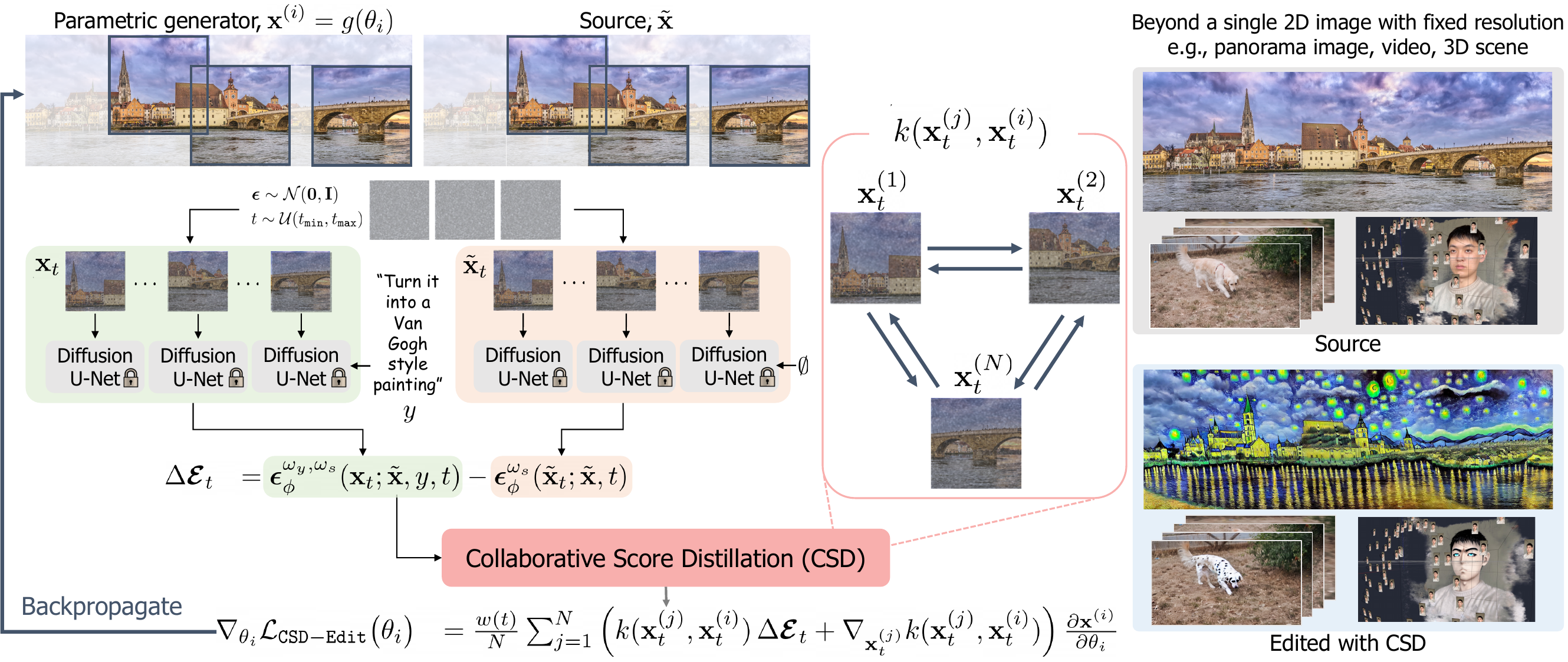}
% example-image-duck}

% \vspace{-0.17in}
\caption{
\textbf{Method overview}. \ssname enables various visual-to-visual translations with two novel components. 
First, a new score distillation scheme using Stein variational gradient descent, which considers inter-sample relationships~(Section~\ref{sec:csd}) to synthesize a set of images while preserving modality-specific consistency constraints.
Second, our method edits images with minimal information given from text instruction by subtracting image-conditional noise estimate instead of random noise during score distillation~(Section~\ref{sec:csdedit}). 
By doing so, \ssname is used for text-guided manipulation of various visual domains, e.g., panorama images, videos, and 3D scenes~(Section~\ref{sec:appl}). 
}
\label{fig:concept_figure}
\vspace{-10pt}
\end{figure*}

Text-to-image diffusion models~\citep{saharia2022photorealistic, ramesh2022hierarchical, nichol2021glide, rombach2022high} have been scaled up by using billions of image-text pairs~\citep{schuhmann2021laion, schuhmann2022laion} and efficient architectures~\citep{ho2020denoising, song2020score, song2020denoising, rombach2022high}, showing impressive capability in synthesizing high-quality, realistic, and diverse images with the text given as an input.
Furthermore, they have branched into various applications, such as image-to-image translation~\citep{meng2021sdedit, bar2022text2live, hertz2022prompt, kawar2022imagic, brooks2022instructpix2pix, mokady2022null,voynov2022sketch}, controllable generation~\citep{zhang2023adding}, or personalization~\citep{gal2022image, ruiz2022dreambooth}. One of the latest applications in this regard is to translate the capability into other complex modalities, {viz.}, beyond 2D images \citep{ho2022imagen, esser2023structure} without modifying diffusion models using modality-specific training data.
This paper focus on the problem of adapting the knowledge of pre-trained text-to-image diffusion models to more complex high-dimensional visual generative tasks beyond 2D images without modifying diffusion models using modality-specific training data.

\begin{figure*}[t]
\centering\small
\includegraphics[width=1\textwidth]{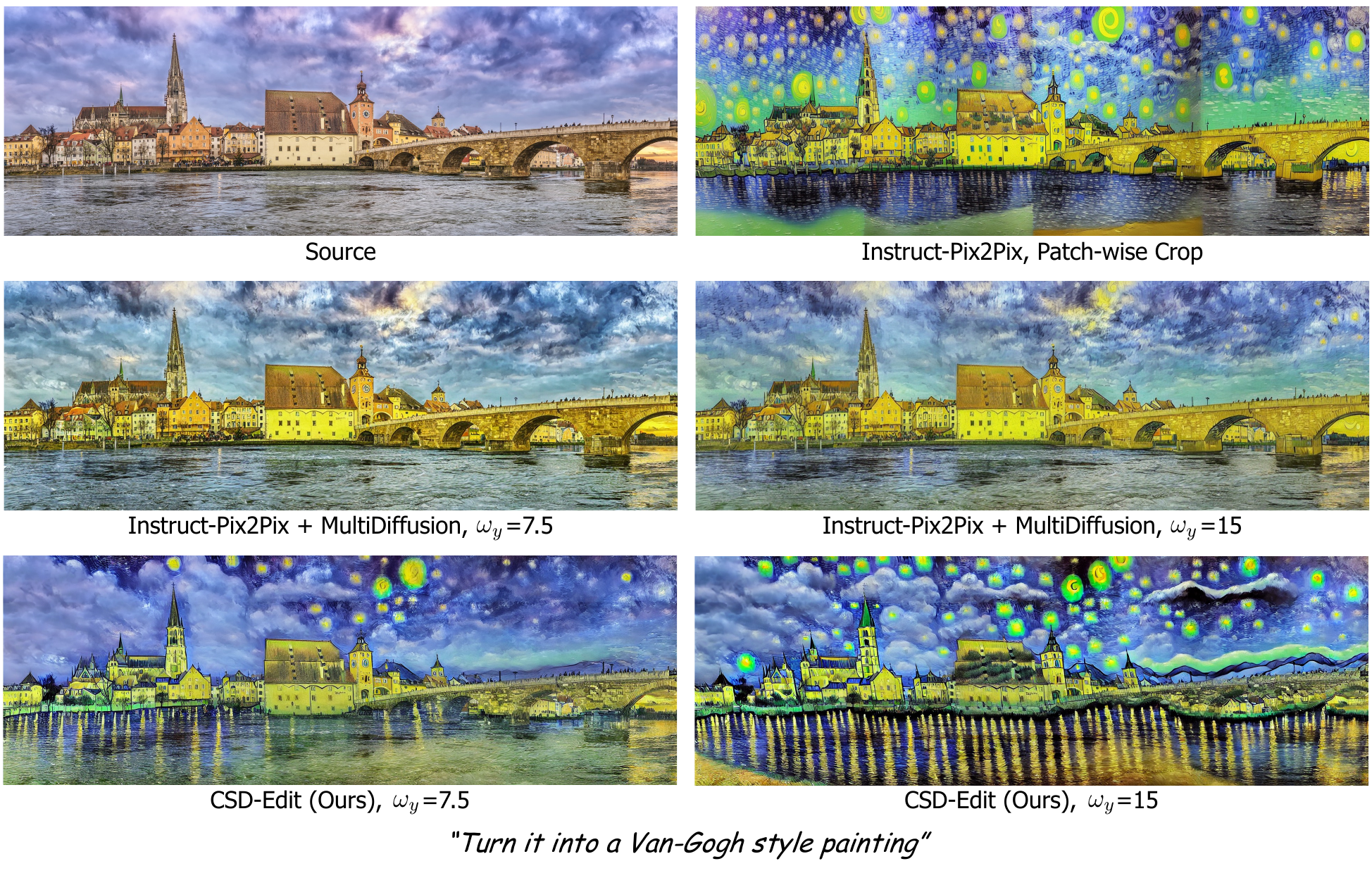}
% \vspace{-0.17in}
\caption{\textbf{Panorama image editing}.
(Top right) Instruct-Pix2Pix~\citep{brooks2022instructpix2pix} on cropped patches results in inconsistent image editing.
(Second row) Instruct-Pix2Pix with MultiDiffusion~\citep{bar2023multidiffusion} edits to consistent image, but less fidelity to the instruction, even with high guidance scale $\omega_y$.
(Third row) \ssname provides consistent image editing with better instruction-fidelity by setting proper guidance scale. 
}
\label{fig:high_res}
\vspace{-10pt}
\end{figure*}

We start from an intuition that many complex visual data, e.g., videos and 3D scenes, are represented as a \emph{set of images} constrained by modality-specific consistency. 
For example, a video is a set of frames requiring temporal consistency, and a 3D scene is a set of multi-view frames with view consistency. Unfortunately, image diffusion models do not have a built-in capability to ensure consistency between a set of images for synthesis or editing because 
their generative sampling process does not take into account the consistency when using the image diffusion model as is.
As such, when applying image diffusion models on these complex data without consistency in consideration, it results in a highly incoherent output, as in Figure~\ref{fig:high_res} (Patch-wise Crop), where one can easily identify where images are stitched.
Such behaviors are also reported in video editing, thus, recent works~\cite{qi2023fatezero,khachatryan2023text2video, liu2023video,ceylan2023pix2video} propose to handle video-specific temporal consistency when using the image diffusion model.

Here, we take attention to an alternative approach, Score Distillation Sampling (SDS)~\citep{poole2022dreamfusion}, which enables the optimization of arbitrary differentiable operators 
by leveraging the rich generative prior of text-to-image diffusion models.  
SDS poses generative sampling as an optimization problem by distilling the learned diffusion density scores.
While \citet{poole2022dreamfusion} has shown the effectiveness of SDS in generating 3D objects from the text by resorting on Neural Radience Fields~\citep{mildenhall20nerf} priors which inherently suppose coherent geometry in 3D space by density modeling, it has not been studied for consistent visual synthesis of other modalities.

In this paper, we propose \emph{Collaborative Score Distillation}~(CSD), a simple yet effective method that extends the singular of the text-to-image diffusion model for consistent visual synthesis. 
The crux of our method is two-fold: first, we establish a generalization of SDS by using Stein variational gradient descent (SVGD), where multiple samples share their knowledge distilled from diffusion models to accomplish inter-sample consistency.
Second, we present \ssname, an effective method for consistent visual editing by leveraging CSD with Instruct-Pix2Pix~\citep{brooks2022instructpix2pix}, a recently proposed instruction-guided image diffusion model (See Figure~\ref{fig:concept_figure}).

We demonstrate the versatility of our method in various applications such as panorama image editing, video editing, and reconstructed 3D scene editing.
In editing a panorama image, we show that \ssname obtains spatially consistent image editing by optimizing multiple patches of an image. 
Also, compared to other methods, our approach achieves a better trade-off between source-target image consistency and instruction fidelity. 
In video editing experiments, \ssname obtains temporal consistency by taking multiple frames into optimization, resulting in temporal frame-consistent video editing. 
Furthermore, we apply \ssname to 3D scene editing and generation, by encouraging consistency among multiple views. 
\section{Preliminaries}\label{sec:prelim}

\subsection{Diffusion models}
Generative modeling with diffusion models consists of a forward process $q$ that gradually adds Gaussian noise to the input $\mathbf{x}_0\sim p_{\tt{data}}(\mathbf{x})$, and a reverse process $p$ which gradually denoises from the Gaussian noise $\mathbf{x}_T\sim \mathcal{N}(\mathbf{0},\mathbf{I})$. 
Formally, the forward process $q(\mathbf{x}_t | \mathbf{x}_{0})$ at timestep $t$ is given by $q(\mathbf{x}_t|\mathbf{x}_0) =\mathcal{N}(\mathbf{x}_t ; \alpha_t \mathbf{x}_0, \sigma_t^2\mathbf{I})$, where $\sigma_t$ and $\alpha_t^2=1-\sigma_t^2$ are pre-defined constants designed for effective modeling~\citep{song2020score, kingma2021variational, karras2022elucidating}.  
Given enough timesteps, reverse process $p$ also becomes a Gaussian and 
the transitions are given by posterior $q$ with optimal MSE denoiser~\citep{pmlr-v37-sohl-dickstein15}, i.e., $p_\phi(\mathbf{x}_{t-1}|\mathbf{x}_t) = \mathcal{N}(\mathbf{x}_{t-1} ; \mathbf{x}_t -\hat{\mathbf{x}}_\phi(\mathbf{x}_t;t), \sigma_t^2 \mathbf{I})$, where $\hat{\mathbf{x}}_\phi(\mathbf{x}_t;t)$ is a learned optimal MSE denoiser. 
\citet{ho2020denoising} proposed to train an U-Net~\citep{ronneberger2015u} autoencoder $\boldsymbol{\epsilon}_\phi(\mathbf{x}_t;t)$ by minimizing following objective:
\begin{align}\label{eq:diffloss}
    \mathcal{L}_{\tt{Diff}}(\phi;\mathbf{x}) = \mathbb{E}_{t\sim \mathcal{U}(0,1), \boldsymbol{\epsilon}\sim\mathcal{N}(\mathbf{0},\mathbf{I})}
    \big[w(t) \| \boldsymbol{\epsilon}_\phi(\mathbf{x}_t;t) -\boldsymbol{\epsilon} \|_2^2\big], \quad \mathbf{x}_t = \alpha_t \mathbf{x}_0 + \alpha_t\boldsymbol{\epsilon}
\end{align}
where $w(t)$ is a weighting function for each timestep $t$. 
Text-to-image diffusion models~\citep{saharia2022photorealistic, ramesh2022hierarchical, rombach2022high,nichol2021glide} are trained by Eq.~\eqref{eq:diffloss} with $\boldsymbol{\epsilon}_\phi(\mathbf{x}_t; y,t)$ that estimates the noise conditioned on the text prompt $y$. 
At inference, those methods rely on Classifier-free Guidance~(CFG)~\citep{ho2022classifier}, which allows higher quality sample generation by introducing additional parameter $\omega_y\geq 1$ as follows:
\begin{align}\label{eq:CFG}
    \boldsymbol{\epsilon}_\phi^\omega (\mathbf{x}_t;y,t) =
    \boldsymbol{\epsilon}_\phi(\mathbf{x}_t; t) +\omega_y \big(\boldsymbol{\epsilon}_\phi(\mathbf{x}_t;y,t)-\boldsymbol{\epsilon}_\phi(\mathbf{x}_t; t)\big)
\end{align}
By setting the appropriate guidance scale $\omega_y>0$, one can improve fidelity to the text prompt at the cost of diversity. Throughout the paper, we refer $p_\phi^{\omega_y}(\mathbf{x}_t;y,t)$ a conditional distribution of a text $y$.

\begin{figure*}[t]
\centering\small
\includegraphics[width=1\textwidth]{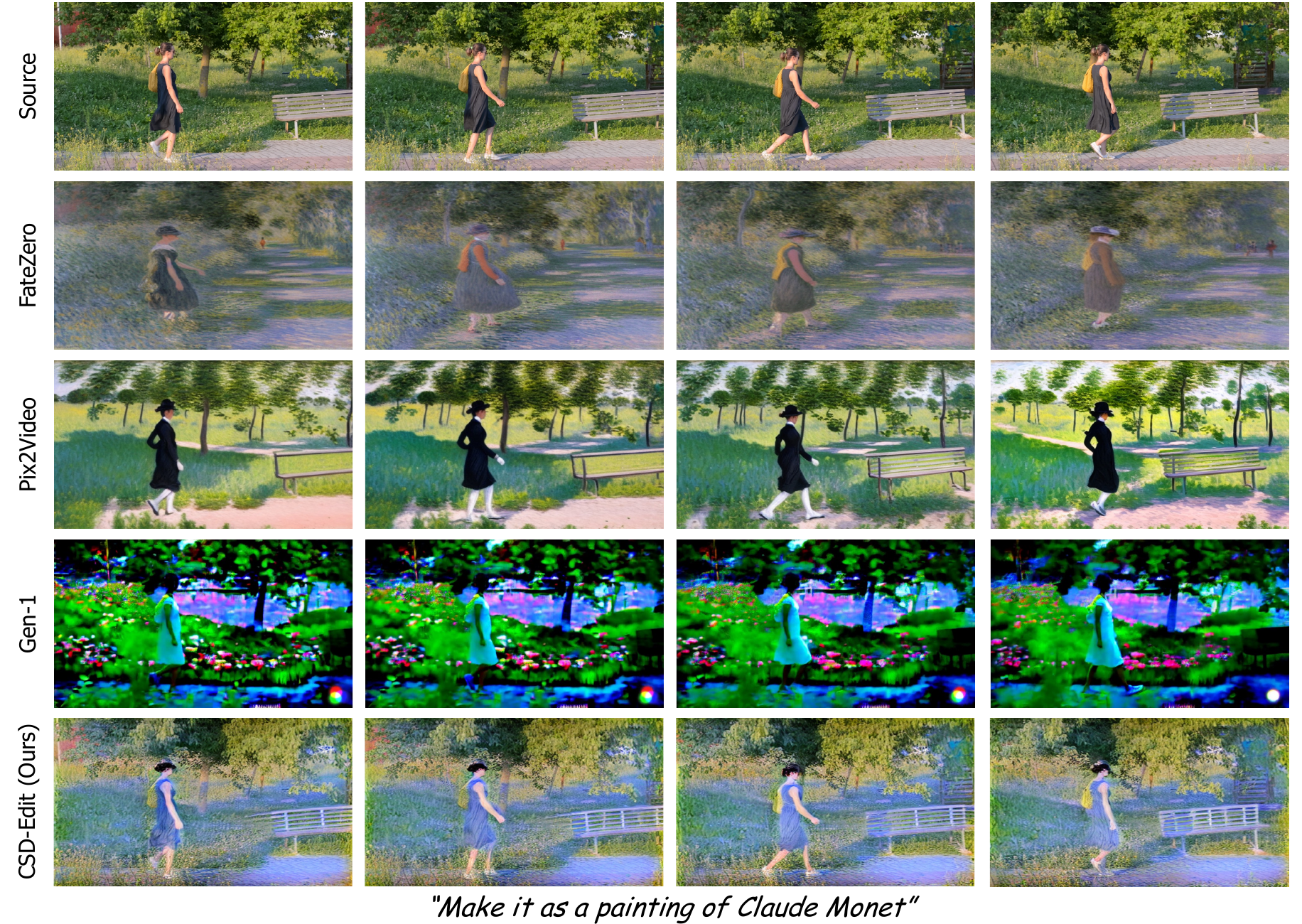}
% \vspace{-0.17in}
\caption{\textbf{Video editing}.
Qualitative results on the lucia video in DAVIS 2017~\citep{Pont-Tuset_arXiv_2017}. \sname shows frame-wise consistent editing providing coherent content across video frames e.g., consistent color and background without changes in person. Compared to Gen-1~\cite{esser2023structure}, a video editing method trained on a large video dataset, \ssname shows high-quality video editing results reflecting given prompts.  
}
\label{fig:video}
\vspace{-12pt}
\end{figure*}

\vspace{0.05in}
{\bf Instruction-based image editing by Instruct-Pix2Pix. }
Recently, many works have demonstrated the capability of diffusion models in editing or stylizing images~\citep{meng2021sdedit, kawar2022imagic,bar2022text2live,hertz2022prompt, brooks2022instructpix2pix}. 
Among them, \citet{brooks2022instructpix2pix} proposed Instruct-Pix2Pix, where they finetuned Stable Diffusion~\citep{rombach2022high} models with the source image, text instruction, edited image (edited by Prompt-to-Prompt~\citep{hertz2022prompt}) triplet to enable instruction-based editing of an image. 
Given source image $\tilde{\mathbf{x}}$ and instruction $y$, the noise estimate at time $t$ is given as 
\begin{align}\label{eq:ip2p}
    \begin{split}
        \boldsymbol{\epsilon}_\phi^{\omega_s,\omega_y}(\mathbf{x}_t;\tilde{\mathbf{x}},y,t) = \boldsymbol{\epsilon}_\phi(\mathbf{x}_t;t) &+
        \omega_s\big( \boldsymbol{\epsilon}_\phi(\mathbf{x}_t;\tilde{\mathbf{x}},t) - \boldsymbol{\epsilon}_\phi(\mathbf{x}_t;t)\big) \\
        &+\omega_y\big(\boldsymbol{\epsilon}_\phi(\mathbf{x}_t;\tilde{\mathbf{x}},y,t) - \boldsymbol{\epsilon}_\phi(\mathbf{x}_t;\tilde{\mathbf{x}},t)\big),
    \end{split}
\end{align}
where $\omega_y$ is CFG parameter for text as in Eq.~\eqref{eq:CFG} and $\omega_s$ is an additional CFG parameter that controls the fidelity to the source image $\tilde{\mathbf{x}}$. 

\subsection{Score distillation sampling}
\citet{poole2022dreamfusion} proposed Score Distillation Sampling~(SDS), an alternative sample generation method by distilling the rich knowledge of text-to-image diffusion models. SDS allows optimization of any differentiable image generator, e.g., Neural Radiance Fields~\citep{mildenhall20nerf} or the image space itself.
Formally, let $\mathbf{x}=g(\theta)$ be an image rendered by a differentiable generator $g$ with parameter $\theta$, then SDS minimizes density distillation loss~\citep{oord2018parallel} which is KL divergence between the posterior of $\mathbf{x} = g(\theta)$ and the text-conditional density $p_\phi^{\omega}$:
\begin{align}\label{eq:KL}
    \mathcal{L}_{\tt{Distill}}\big(\theta; \mathbf{x}=g(\theta)\big) = \mathbb{E}_{t,\boldsymbol{\epsilon}}\big[ \alpha_t/\sigma_t\,D_{\tt{KL}}\big( q\big(\mathbf{x}_t|\mathbf{x}=g(\theta)\big) \,\|\, p_\phi^{\omega}(\mathbf{x}_t; y,t)\big) \big].
\end{align}
For an efficient implementation, SDS updates the parameter $\theta$ by randomly choosing timesteps $t\sim \mathcal{U}(t_{\tt{min}}, t_{\tt{max}})$ and forward $\mathbf{x}=g(\theta)$ with noise $\boldsymbol\epsilon\sim \mathcal{N}(\mathbf{0},\mathbf{I})$ to compute the gradient as follows: 
\begin{align}\label{eq:sds}
    \nabla_\theta \mathcal{L}_{\tt{SDS}}\big(\theta; \mathbf{x}=g(\theta)\big) = \mathbb{E}_{t,\boldsymbol{\epsilon}}\left[ w(t) \big(\boldsymbol{\epsilon}_\phi^{\omega}(\mathbf{x}_t; y,t) - \boldsymbol{\epsilon}\big)\frac{\partial \mathbf{x}}{\partial \theta} \right]. 
\end{align}
Remark that the U-Net Jacobian $\partial\epsilon_\phi^{\omega}(\mathbf{z}_t;y,t)/\partial \mathbf{z}_t$ is omitted as it is computationally expensive to compute, and degrades performance when conditioned on small noise levels.
The range of timesteps $t_{\tt{min}}$ and $t_{\tt{max}}$ are chosen to sample from not too small or large noise levels, and the guidance scales are chosen to be larger than those used for image generation.

\subsection{Stein variational gradient descent}\label{sec:svgd}
The original motivation of Stein variational gradient descent (SVGD)~\citep{liu2016stein} is to solve a variational inference problem, where the goal is to approximate a target distribution from a simpler distribution by minimizing KL divergence.
Formally, suppose $p$ is a target distribution with a known score function
$\nabla_{\mathbf{x}} \log p(\mathbf{x})$ that we aim to approximate, and $q(\mathbf{x})$ is a known source distribution.
\citet{liu2016stein} showed that the steepest descent of KL divergence between $q$ and $p$ is given as follows:
\begin{align}\label{eq:steindis}
    \mathbb{E}_{q(\mathbf{x})}\big[\mathbf{f}(\mathbf{x})^\top \nabla_\mathbf{x}\log p(\mathbf{x}) + \text{Tr}(\nabla_\mathbf{x} \mathbf{f}(\mathbf{x}))\big],
\end{align}
where $\mathbf{f}:\mathbb{R}^D\rightarrow\mathbb{R}^D$ is any smooth vector function that satisfies $\lim_{\|\mathbf{x}\|\rightarrow\infty}p(\mathbf{x})\mathbf{f}(\mathbf{x}) = 0$.
Remark that Eq.~\eqref{eq:steindis} becomes zero if we replace $q(\mathbf{x})$ with $p(\mathbf{x})$ in the expectation term, which is known as Stein's identity~\cite{gorham2017measuring}.
Here, the choice of the critic $\mathbf{f}$ is crucial in its convergence and computational tractability. 
To that end, \citet{liu2016stein} proposed to constrain $\mathbf{f}$ in the Reproducing Kernel Hilbert Space (RKHS) which yields a closed-form solution. 
Specifically, given a positive definite kernel $k:\mathbb{R}^D\times\mathbb{R}^D\rightarrow\mathbb{R}^+$, Stein variational gradient descent provides the greedy directions as follows:
\begin{align}\label{eq:svgd}
    \mathbf{x}\leftarrow \mathbf{x} -\eta\Delta\mathbf{x}, \quad \Delta\mathbf{x} = \mathbb{E}_{q(\mathbf{x}^\prime)}\big[k(\mathbf{x}, \mathbf{x}^\prime) \nabla_{\mathbf{x}^\prime} \log p(\mathbf{x}^\prime) + \nabla_{\mathbf{x}^\prime}k(\mathbf{x},\mathbf{x}^\prime)\big],
\end{align}
with small step size $\eta>0$. The SVGD update in Eq.~~\eqref{eq:svgd} consists of two terms that play different roles: the first term moves the particles towards the high-density region of target density $p(\mathbf{x})$, where the direction is smoothed by kernels of other particles. 
The second term acts as a repulsive force that prevents the mode collapse of particles. 
One can choose different kernel functions, while we resort to standard Radial Basis Function (RBF) kernel $k(\mathbf{x},\mathbf{x}^\prime) = \exp(-\frac{1}{h}\|\mathbf{x}-\mathbf{x}^\prime\|_2^2)$ with bandwidth $h>0$. 
\section{Method}

\begin{figure*}[t]
\centering\small
\includegraphics[width=1\textwidth]{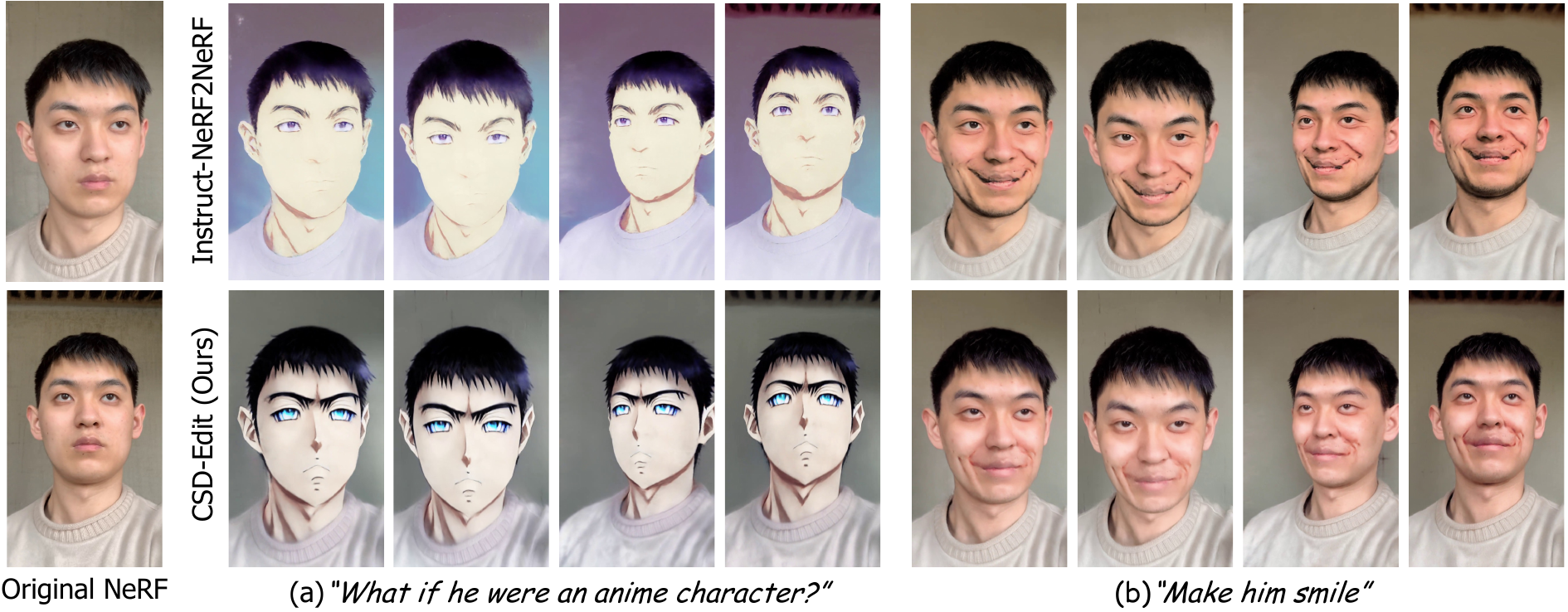}
% \vspace{-0.17in}
\caption{\textbf{3D NeRF scene editing}.
Visualizing novel-views of edited Fangzhou NeRF scene~\citep{wang2022nerf}. 
% \ssname can properly edit the 3D scenes following the given prompts. 
\ssname leads to high-quality editing of 3D scenes and better preserves semantics of source scenes, e.g., obtains sharp facial details (left) and makes him smile without giving beard (right). 
}
\label{fig:scene_3d}
\vspace{-10pt}
\end{figure*}

In this section, we introduce \emph{Collaborative Score Distillation}~(\sname) for consistent synthesis and editing of multiple samples. 
We first derive a collaborative score distillation method using Stein variational gradient descent (Section~\ref{sec:csd}) and propose an effective image editing method using \sname, i.e., \ssname, that leads to coherent editing of multiple images with instruction~(Section~\ref{sec:csdedit}).
Lastly, we present various applications of \ssname in editing panorama images, videos, and 3D scenes~(Section~\ref{sec:appl}).

\subsection{Collaborative score distillation}\label{sec:csd}
Suppose a set of parameters $\{\theta_i\}_{i=1}^N$ that generates images $\mathbf{x}^{(i)} = g(\theta_i)$. Similar to SDS, our goal is to update each $\theta_i$ by distilling the smoothed densities from the diffusion model by minimizing KL divergence in Eq.~\eqref{eq:KL}. 
On the contrary, {\sname} solves Eq.~\eqref{eq:KL} using SVGD demonstrated in Section~\ref{sec:svgd} so that each $\theta_{i}$ can be updated in sync with updates of other parameters in the set $\{\theta_{i}\}_{i=1}^{N}$.
At each update, \sname samples $t\sim\mathcal{U}(t_{\tt{min}},t_{\tt{max}})$ and $\boldsymbol{\epsilon}\sim\mathcal{N}(\mathbf{0}, \mathbf{I})$, and update each $\theta_i$ as follows:
% In contrast to SDS which updates each parameter independently, we update parameters synchronously via SVGD as follows: 
\begin{align}\label{eq:csd-emp}
    \nabla_{\theta_i} \mathcal{L}_{\tt{CSD}}\big(\theta_i\big) = 
    \frac{w(t)}{N}\sum_{j=1}^N \left(k(\mathbf{x}_t^{(j)}, \mathbf{x}_t^{(i)})(\boldsymbol{\epsilon}_\phi^{\omega}(\mathbf{x}_t^{(j)};y,t) - \boldsymbol{\epsilon} ) + \nabla_{\mathbf{x}_t^{(j)}} k(\mathbf{x}_t^{(j)}, \mathbf{x}_t^{(i)})\right)\frac{\partial \mathbf{x}^{(i)}}{\partial \theta_i},
\end{align}
for each $i=1,2,\ldots, N$. We refer to Appendix~\ref{appen:derivation} for full derivation. 
Note \sname is equivalent to SDS in Eq.~\eqref{eq:sds} when $N=1$, showing that \sname is a generalization of SDS to multiple samples. As the pairwise kernel values are multiplied by the noise prediction term, each parameter update on $\theta_i$ is affected by other parameters, i.e., the scores are mixed with importance weights according to the affinity among samples.
The more similar samples tend to exchange more score updates, while different samples tend to interchange the score information less. The gradient of the kernels acts as a repulsive force that prevents the mode collapse of samples. 
Moreover, we note that Eq.~\eqref{eq:csd-emp} does not make any assumption on the relation between $\theta_i$'s or their order besides them being a set of images to be synthesized coherently with each other. 
As such, \sname is also applicable to arbitrary image generators, as well as text-to-3D synthesis in DreamFusion~\citep{poole2022dreamfusion}, which we compare in Section~\ref{sec:ablation}.

\subsection{Text-guided editing by collaborative score distillation}\label{sec:csdedit}
In this section, we introduce a text-guided visual editing method using Collaborative Score Distillation (\ssname).
Given source images $\tilde{\mathbf{x}}^{(i)} \,{=}\, g(\tilde{\theta}_i)$ with parameters $\tilde{\theta}_i$, we optimize new target parameters $\{\theta_i\}_{i=1}^N$ with $\mathbf{x}^{(i)}\,{=}\,g(\theta_i)$ such that 1) each $\mathbf{x}^{(i)}$ follows the instruction prompt, 2) preserves the semantics of source images as much as possible, and 3) the obtained images are consistent with each other. 
To accomplish these, we update each parameter $\theta_i$, initialized with $\tilde{\theta}_i$, using {\sname} with noise estimate $\boldsymbol{\epsilon}_\phi^{\omega_y,\omega_s}$ of Instruct-Pix2Pix. However, this approach often results in blurred outputs, leading to the loss of details of the source image (see Figure~\ref{fig:abl}). 
This is because the score distillation term subtracts random noise $\boldsymbol{\epsilon}$, which perturbs the undesirable details of source images. 

We handle this issue by adjusting the noise prediction term that enhances the consistency between source and target images.
Subtracting a random noise $\boldsymbol{\epsilon}$ in Eq.~\eqref{eq:sds} when computing the gradient is a crucial factor, which helps optimization by reducing the variance of a gradient.
Therefore, we amend the optimization by changing the random noise into a better baseline function. 
Since our goal is to edit an image with only minimal information given text instructions, we set the baseline by the image-conditional noise estimate of the Instruct-Pix2Pix model without giving text instructions on the source image. 
To be specific, our \ssname is given as follows:
\begin{align}\label{eq:csd-edit}
    \begin{split}
        \nabla_{\theta_i} \mathcal{L}_{\tt{CSD-Edit}}\big(\theta_i\big) &= \frac{w(t)}{N}\sum_{j=1}^N \left(k(\mathbf{x}_t^{(j)}, \mathbf{x}_t^{(i)})\,\Delta \boldsymbol{\mathcal{E}}_t^{(i)} + \nabla_{\mathbf{x}_t^{(j)}} k(\mathbf{x}_t^{(j)}, \mathbf{x}_t^{(i)})\right)\frac{\partial \mathbf{x}^{(i)}}{\partial \theta_i},\\
        \Delta \boldsymbol{\mathcal{E}}_t^{(i)} &= \boldsymbol{\epsilon}_\phi^{\omega_y,\omega_s}(\mathbf{x}_t^{(i)};\tilde{\mathbf{x}}, y,t) - \boldsymbol{\epsilon}_\phi^{\omega_s}(\tilde{\mathbf{x}}_t^{(i)};\tilde{\mathbf{x}}, t).
    \end{split}
\end{align}
In Section~\ref{sec:ablation}, we validate our findings on the effect of baseline noise on image editing performance. 
We notice that \ssname presents an alternative way to utilize Instruct-Pix2Pix in image-editing without any finetuning of diffusion models, by posing an optimization problem.

\subsection{\ssname for various complex visual domains}\label{sec:appl}
\paragraph{Panorama image editing.}
Diffusion models are usually trained on a fixed resolution (e.g., 512$\times$512 for Stable Diffusion~\citep{rombach2022high}), thus when editing a panorama image~(i.e., an image with a large aspect ratio), the editing quality significantly degrades.
Otherwise, one can crop an image into smaller patches and apply image editing on each patch. 
However this results in spatially inconsistent images (see Figure~\ref{fig:high_res}, Patch-wise Crop, Appendix~\ref{appen:vis}). 
To that end, we propose to apply \ssname on patches to obtain spatially consistent editing of an image, while preserving the semantics of source image.
Following \citep{bar2023multidiffusion}, we sample patches of size 512$\times$512 that overlap using small stride and apply \ssname on the latent space of Stable Diffusion~\citep{rombach2022high}.
Since we allow overlapping, some pixels might be updated more frequently. Thus, we normalize the gradient of each pixel by counting the appearance.

\paragraph{Video editing.}
Editing a video with an instruction should satisfy the following: 1) temporal consistency between frames such that the degree of changes compared to the source video should be consistent across frames, 2) ensuring that desired edits in each edited frame are in line with the given prompts while preserving the original structure of source video, and 3) maintaining the sample quality in each frame after editing.
To meet these requirements, we randomly sample a batch of frames and update them with \ssname to achieve temporal consistency between frames. 

\paragraph{3D scene editing.}
We consider editing a 3D scene reconstructed by a Neural Radiance Field~(NeRF)~\citep{mildenhall20nerf}, which represents volumetric 3D scenes using 2D images. 
To edit reconstructed 3D NeRF scenes, it is straightforward to update the training views with edited views and finetune the NeRF with edited views. 
Here, the multi-view consistency between edited views should be considered since inconsistencies between edits across multiple viewpoints lead to blurry and undesirable artifacts, hindering the optimization of NeRF. 
To mitigate this, \citet{instructnerf2023} proposed Instruct-NeRF2NeRF, which performs editing on a subset of training views and updates them sequentially at training iteration with intervals. 
However, image-wise editing results in inconsistencies between views, thus they rely on the ability of NeRF in achieving multi-view consistency.
Contrary to Instruct-NeRF2NeRF, we update the dataset with multiple consistent views through \ssname, which serves as better training resources for NeRF, leading to less artifacts and better preservation of source 3D scene.

\section{Experiments}
\begin{table*}[t]
\centering\small
\vspace{-0.1in}

\begin{minipage}{.45\textwidth}
\centering\small
\includegraphics[width=0.9\textwidth]{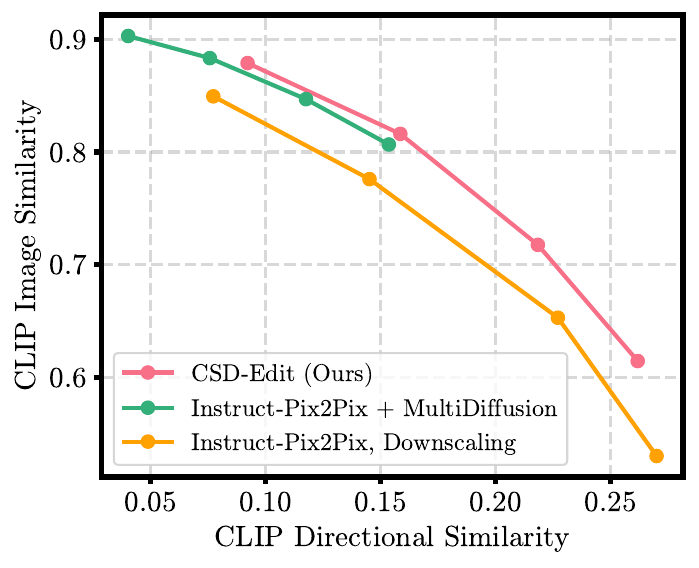}
\vspace{-0.1in}
\captionof{figure}{\textbf{Panorama image editing}.
Comparison of \ssname with baselines at different guidance scales $\omega_y\in\{3.0, 5.0, 7.5, 10.0\}$.
}\label{fig:exp_panorama}
\end{minipage}
\hfill
\begin{minipage}{.5\textwidth}
{
\centering\Large

\caption{\textbf{Video editing}. Quantitative comparison of \ssname with baselines on video editing. Bold indicates the best results.
}\label{tab:main_video}
\vspace{-0.1in}
\resizebox{\textwidth}{!}{
        \begin{tabular}{c|c|c|c}
            \toprule 
             & {CLIP Directional} & {CLIP Image} & {LPIPS}\\ 
             & {Similarity} $\uparrow$ & {Consistency} $\uparrow$ & $\downarrow$\\ 
             \midrule
             FateZero~\cite{qi2023fatezero} & 0.314 & 0.948 & 0.267\\
             Pix2Vid~\cite{ceylan2023pix2video} & 0.230 & 0.949 & 0.283\\
             \rowcolor{Gray}\textbf{\ssname (Ours)} & \textbf{0.320} & \textbf{0.957} & \textbf{0.236}\\
             \bottomrule
        \end{tabular}
}

\vspace{0.15in}

\caption{\textbf{3D scene editing}. Quantitative  comparison of \ssname with baselines on 3D scene editing. Bold indicates the best results.
}\label{tab:main_nerf}
\vspace{-0.1in}
\resizebox{\textwidth}{!}{
        \begin{tabular}{c|c|c|c}
            \toprule 
             & {CLIP Directional} & {CLIP Image} & {LPIPS}\\ 
             & {Similarity} $\uparrow$ & {Consistency} $\uparrow$ & $\downarrow$\\ 
             \midrule
             IN2N~\cite{brooks2022instructpix2pix} & 0.230 & 0.994 & 0.048\\
             \rowcolor{Gray}\textbf{\ssname (Ours)} & \textbf{0.239} & \textbf{0.995} & \textbf{0.043}\\
             \bottomrule
        \end{tabular}
}

}
\end{minipage}

\vspace{-10pt}
\end{table*}

\subsection{Text-guided panorama image editing}\label{sec:exp-panorama}
For the panorama image-to-image translation task, we compare \ssname with different versions of Instruct-Pix2Pix: one is which using naive downsizing to $512\times512$ and performing Instruct-Pix2Pix, and another is updating Instruct-Pix2Pix on the patches as in MultiDiffusion~\citep{bar2023multidiffusion} (Instruct-Pix2Pix + MultiDiffusion). 
For comparison, we collect a set of panorama images (i.e., which aspect ratio is higher than 3), and edit each image to various artistic styles and different guidance scales $\omega_y$.
For evaluation, we use pre-trained CLIP~\citep{radford2021learning} to measure two different metrics: 1) consistency between source and target images by computing similarity between two image embeddings, and 2) CLIP directional similarity~\citep{gal2022stylegan} which measures how the change in text agrees with the change in the images. The experimental details are in Appendix~\ref{appen:high_res}. 

In Figure~\ref{fig:exp_panorama}, we plot the CLIP scores of different image editing methods with different guidance scales. 
We notice that \ssname provides the best trade-off between the consistency between source and target images and fidelity to the instruction.
Figure~\ref{fig:high_res} provides a qualitative comparison between panorama image editing methods. 
Remark that Instruct-Pix2Pix + MultiDiffusion is able to generate spatially consistent images, however, the edited images show inferior fidelity to the text instruction even when using a large guidance scale. Additional qualitative results are in Appendix~\ref{appen:vis}.

\subsection{Text-guided video editing}\label{sec:exp-video}
For the video editing experiments, we primarily compare \ssname with existing zero-shot video editing schemes that employ text-to-image diffusion models such as FateZero~\citep{qi2023fatezero}, and Pix2Video~\cite{ceylan2023pix2video}. 
To emphasize the effectiveness of \ssname against learning-based schemes, we also compare it with Gen-1~\citep{esser2023structure}, a state-of-the-art video editing method trained on a large-scale video dataset. For quantitative evaluation, we report CLIP image-text directional similarity as in Section~\ref{sec:exp-panorama} to measure alignment between changes in texts and images. Also, we measure CLIP image consistency and LPIPS~\citep{zhang2018perceptual} between consecutive frames to evaluate temporal consistency. 
We utilize video sequences from the popular DAVIS~\citep{Pont-Tuset_arXiv_2017} dataset at a resolution of $1920\times 1080$. Please refer to Appendix~\ref{appen:video} for a detailed description of the baseline methods and experimental setup. 

Table~\ref{tab:main_video} summarize quantitative comparison between \ssname and the baselines.
We notice that \ssname consistently outperforms the existing zero-shot video editing schemes in terms of both temporal consistency and fidelity to given text prompts.
Moreover, Figure~\ref{fig:video} qualitatively demonstrate the superiority of \sname over the baselines on video-stylization and object-aware editing tasks. Impressively, \sname shows comparable editing performance to Gen-1 even without training on a large-scale video dataset and any architectural modification to the diffusion model. Additional qualitative results are in Appendix~\ref{appen:vis}. 

\subsection{Text-guided 3D scene editing}
For the text-guided 3D scene editing experiments, we mainly compare our approach with Instuct-NeRF2NeRF~(IN2N)~\citep{instructnerf2023}. 
For a fair comparison, we exactly follow the experimental setup which they used, and faithfully find the hyperparameters to reproduce their results. 
For evaluation, we render images at the novel views (i.e., views not seen during training), and report CLIP image similarity and LPIPS between consecutive frames in rendered videos to measure multi-view consistency, as well as CLIP image-text similarity to measure fidelity to the instruction. 
Detailed explanations for each dataset sequence and training details can be found in Appendix~\ref{appen:3d}.

Figure~\ref{fig:scene_3d} and Table~\ref{tab:main_nerf} summarize the comparison between \ssname and IN2N.
We notice that \ssname enables a wide-range control of 3D NeRF scenes, such as delicate attribute manipulation (e.g., facial expression alterations) and scene-stylization~(e.g., conversion to the animation style). 
Especially, we notice two advantages of \ssname compared to IN2N. 
First, \ssname presents high-quality details to the edited 3D scene by providing multi-view consistent training views during NeRF optimization. 
In Figure~\ref{fig:scene_3d}, one can observe that \ssname captures sharp details of anime character, while IN2N results in blurry face. 
Second, \ssname is better at preserving the semantics of source 3D scenes, e.g., backgrounds or colors. For instance in Figure~\ref{fig:scene_3d}, we notice that \ssname allows subtle changes in facial expressions without changing the color of the background or adding a beard to the face. 

\begin{table*}[t]
\centering\small
\vspace{-0.07in}
\centering\small
\includegraphics[width=\textwidth]{{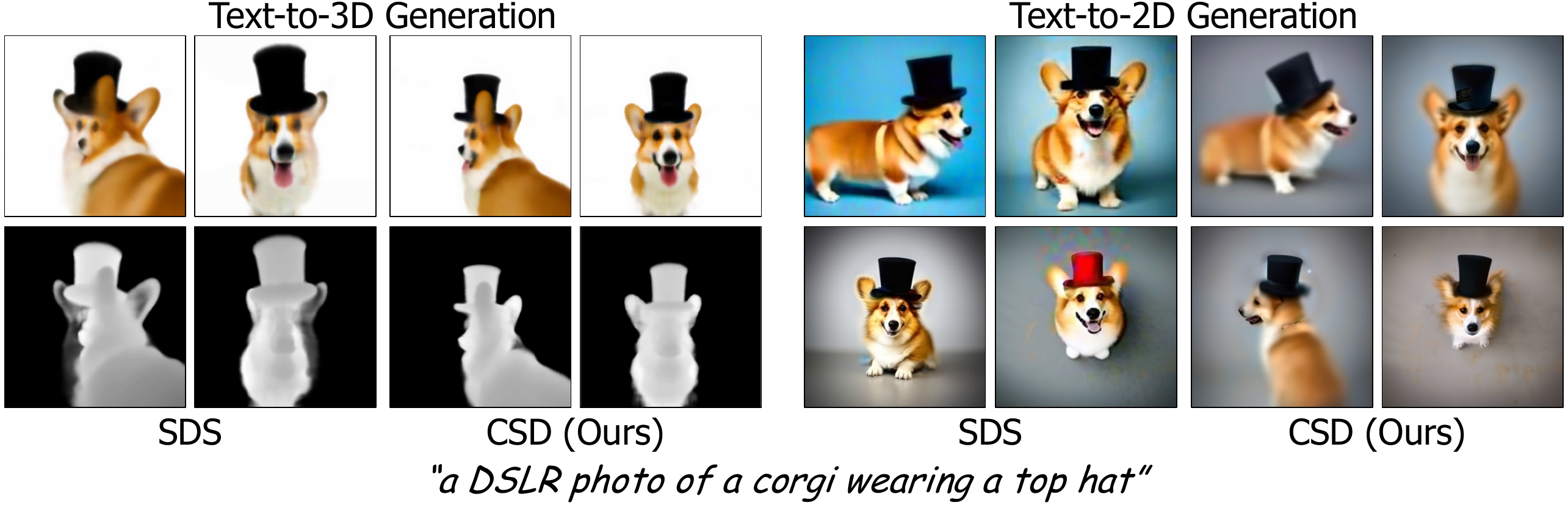}}
\captionof{figure}{\textbf{Text-to-3D generation}. (Left) \sname helps capturing coherent geometry in synthesizing 3D object. (Right) \sname generates coherent images conditioned on view-dependent prompts. 
}\label{fig:gen_3d}
\vspace{-10pt}
\end{table*}
\subsection{Ablation study}\label{sec:ablation}
\paragraph{\sname for text-to-3D generation.} 
We explore the effectiveness of \sname in text-to-3D generation tasks following DreamFusion~\citep{poole2022dreamfusion}. 
We train a coordinate MLP-based NeRF architecture from scratch using text-to-image diffusion models. 
Since the pixel-space diffusion model that DreamFusion used~\citep{poole2022dreamfusion} is not publicly available, we used an open-source implementation of pixel-space text-to-image diffusion model.\footnote{\url{https://github.com/deep-floyd/IF}}
When using \sname for text-to-3D generation, we empirically observe that using LPIPS~\citep{zhang2018unreasonable} as a distance for RBF kernel worked well.  
We refer to Appendix~\ref{appen:add-dreamfusion} for details.

Given a set of text prompts, we run both DreamFusion and DreamFusion with \sname with a fixed seed.
In Figure~\ref{fig:gen_3d}, we visualize generated examples.
Remark that DreamFusion and DreamFusion + \sname tend to generate similar objects,
but we observe that \sname often adds better details that complement the poor quality of one that made by DreamFusion.
For instance, in Figure~\ref{fig:gen_3d}, \sname removes blurry artifacts in the synthesized 3D NeRF scene, which is often caused by inconsistent view distillation. 
Also in Figure~\ref{fig:gen_3d}, we verify that the \sname generates more coherent images when conditioned on view-dependent prompts which were used in DreamFusion. 
We refer to Appendix~\ref{appen:add-dreamfusion} for more examples of text-to-3D generation.

\begin{figure*}[t]
\centering\small
\includegraphics[width=1\textwidth]{{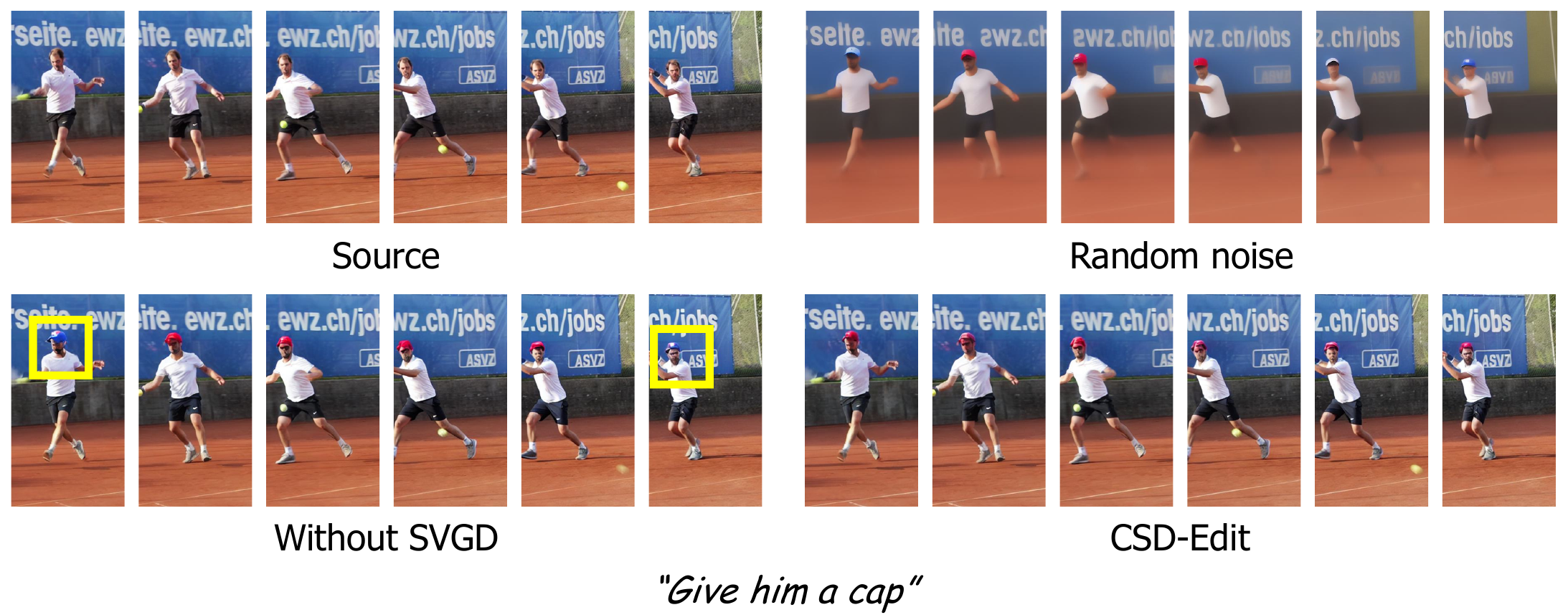}}
\vspace{-10pt}
\captionof{figure}{\textbf{Ablation study}. Given a source video (top left), \ssname without SVGD results in inconsistent frames (bottom left), and subtracting random noise in \ssname results in loss of details (top right).
\ssname obtains consistency between frames without loss of semantics~(bottom right).
% backgrouComparison of edited results on the tennis video in DAVIS 2017~\citep{Pont-Tuset_arXiv_2017}.
}\label{fig:abl}
\vspace{-6pt}
\end{figure*}

\paragraph{Ablation on the components of \sname.}
To demonstrate the effect of our method, we present an ablation study on a video editing experiment. 
To verify the role of communication between samples using SVGD, we compare the editing results with and without SVGD. 
Also, to verify the role of baseline noise in \ssname, we provide result when using random noise as baseline. 
As shown in Figure~\ref{fig:abl}, \ssname consistently edits a source video adding a red cap on a man's head when given the instruction ``give him a cap.'' However, without SVGD, the edits between frames are inconsistent, for example, blue caps or red caps appear both on the edited frames. In addition, if we set the baseline noise as the random noise injected into the source and target image, each frame gets blurry and loses the original structures, e.g., blurred legs and backgrounds.

% \placeholder{4}

\section{Related work}\label{sec:rel}
Following remarkable success of text-to-image diffusion models~\cite{rombach2022high, ho2022imagen, saharia2022photorealistic, ramesh2022hierarchical, balaji2022ediffi}, numerous works have attempted to exploit rich knowledge of text-to-image diffusion models for various visual editing tasks including images~\cite{meng2021sdedit,couairon2022diffedit, kawar2022imagic, valevski2022unitune, brooks2022instructpix2pix,hertz2022prompt,mokady2022null}, videos~\cite{wu2022tune, ceylan2023pix2video}, 3D scenes~\cite{instructnerf2023}, etc. However, extending existing image editing approaches to more complex visual modalities often faces a new challenge; consistency between edits, e.g., spatial consistency in high-resolution images, temporal consistency in videos, and multi-view consistency in 3D scenes. 
While prior works primarily focus on designing task-specific methods~\cite{liu2023video, qi2023fatezero, ceylan2023pix2video} or model fine-tuning for complex modalities~\cite{wu2022tune}, we present a modality-agnostic novel method for editing, effectively capturing consistency between samples.

The most related to our work is DreamFusion~\cite{poole2022dreamfusion}, which introduced Score Distillation Sampling (SDS) for creation of 3D assets, leveraging the power of text-to-image diffusion models. Despite the flexible merit of SDS to enable the optimization of arbitrary differentiable operators, most works mainly focus on applying SDS to enhance the synthesis quality of 3D scenes by introducing 3D specific frameworks~\cite{lin2022magic3d, tsalicoglou2023textmesh, melas2023realfusion, chen2023fantasia3d, tang2023make}. Although there exists some work to apply SDS for visual domains other than 3D scenes, they have limited their scope to image editing~\cite{hertz2023delta}, or image generation~\cite{song2022diffusion}. Here, we clarify that our main focus is not to improve the performance of SDS for a specific task, but rather to shift the focus to generalizing it from a new perspective in a principled way. To the best of our knowledge, we are the first to center our work on the generalization of SDS and introduce a novel method that simply but effectively adapts text-to-image diffusion models to diverse high-dimensional visual syntheses beyond a single 2D image with fixed resolution. 

\section{Conclusion}\label{sec:con}
In this paper, we propose Collaborative Score Distillation (CSD) for consistent visual synthesis and manipulation. CSD is built upon Stein variational gradient descent, where multiple samples share their knowledge distilled from text-to-image diffusion models during the update. 
Furthermore, we propose CSD-Edit that gives us consistent editing of images by distilling minimal, yet sufficient information from instruction-guided diffusion models.
We demonstrate the effectiveness of our method in text-guided translation of diverse visual contents, such as in high-resolution images, videos, and real 3D scenes, outperforming previous methods both quantitatively and qualitatively.

% \vspace{0.05in}
{\bf Limitations. }
Since we use pre-trained text-to-image diffusion models, there are some cases where the results are imperfect due to the inherent inability of diffusion models in understanding language.
Also, our method might be prone to the underlying societal biases in diffusion models. 
See Appendix~\ref{appen:limitation}.

{\bf Societal impact. }
Our method enables consistent editing of visual media. On the other hand, our method is not free from the known issues that text-to-image models carry when used by malicious users.
%Our method could be potentially misused by malicious users who aim to synthesize or edit visual content to agitate viewers. 
We expect future research on the detection of generated visual content.
See Appendix~\ref{appen:broad}.

% \clearpage
{%\small
\bibliographystyle{unsrtnat}  % custom unsrt+abbrv
\bibliography{reference}
}
\newpage
% \section{Appendix}
\newpage
\appendix
\onecolumn
\begin{center}
{\bf {\LARGE Appendix}}
\end{center}
\begin{center}
    \textbf{Website:} \url{https://subin-kim-cv.github.io/CSD}
\end{center}

\section{Technical details}\label{appen:derivation}
In this section, we provide detailed explanations on the proposed methods, \sname and \ssname. 
\paragraph{\sname derivation.}
Consider a set of parameters $\{\theta_i\}_{i=1}^N$ which generates images $\mathbf{x}^{(i)}=g(\theta_i)$. For each timestep $t\sim\mathcal{U}(t_{\tt{min}}, t_{\tt{max}})$, we aim at minimizing the following KL divergence 
\begin{align*}
    D_{\tt{KL}}\big( q(\mathbf{x}_t^{(i)} | \mathbf{x}^{(i)}=g(\theta_i)) \| p_\phi(\mathbf{x}_t;y,t) \big)
\end{align*}
for each $i=1,2,\ldots, N$ via SVGD using Eq.~\eqref{eq:svgd}. 
To this end, we approximate the score function, (i.e., gradient of log-density) by the noise predictor from diffusion model as follows:
$$\nabla_{\mathbf{x}_t^{(i)}} \log p_\phi(\mathbf{x}_t^{(i)} ;y,t) \approx -\frac{\boldsymbol{\epsilon}_\phi(\mathbf{x}_t^{(i)};y,t)}{\sigma_t}.$$
Then, the gradient of score function with respect to parameter $\theta_i$ is given by
\begin{align}\label{eq:deriv}
    \nabla_{\theta_i} \log p_\phi(\mathbf{x}_t^{(i)};y,t) = \nabla_{\mathbf{x}_t^{(i)}} \log p_\phi(\mathbf{x}_t^{(i)} ;y,t) \frac{\partial \mathbf{x}_t^{(i)}}{\partial \theta_i} \approx -\frac{\alpha_t}{\sigma_t} \boldsymbol{\epsilon}_\phi(\mathbf{x}_t^{(i)};y,t) \frac{\partial \mathbf{x}^{(i)}}{\partial\theta},
\end{align}
for each $i=1,\ldots N$. 
Finally, to derive \sname, we plug Eq.~\eqref{eq:deriv} to Eq.~\eqref{eq:svgd} to attain Eq.~\eqref{eq:csd-emp}. 
Also, we subtract the noise $\boldsymbol{\epsilon}$, which helps reducing the variance of gradient for better optimization.
Following DreamFusion~\citep{poole2022dreamfusion}, we do not compute the Jacobian of U-Net. 
At high level, \sname takes the gradient update on each $\mathbf{x}^{(i)}$ using SVGD and update $\theta_i$ by simple chain rule without computing the Jacobian. 
This formulation makes \sname as a straightforward generalization to SDS for multiple samples and leads to effective gradient for optimizing with consistency among batch of samples.

\paragraph{CSD-Edit derivation.}
As mentioned above, we subtract the random noise to reduce the variance of \sname gradient estimation. 
This is in a similar manner to the variance reduction in policy gradient~\citep{schulman2015high}, where having proper baseline function results in faster and more stable optimization. 
Using this analogy, our intuition is build upon that setting better baseline function can ameliorate the optimization of \sname. 
Thus, in image-editing via \ssname, we propose to use image-conditional noise estimate as a baseline function. 
This allows \ssname to optimize the latent driven by only the influence of instruction prompts. 
We notice that similar observations were proposed in Delta Denoising Score~(DDS)~\citep{hertz2023delta}, where they introduced an image-to-image translation method that is based on SDS, and the difference of the noise estimate from target prompt and that from source prompt are used. 
Our \sname can be combined with DDS by changing the noise difference term as follows:
\begin{align*}
    \Delta \boldsymbol{\mathcal{E}}_t = \boldsymbol{\epsilon}_\phi(\mathbf{x}_t; y_{\tt{tgt}},t) - \boldsymbol{\epsilon}_\phi(\tilde{\mathbf{x}}_t; y_{\tt{src}}, t),
\end{align*}
where $\mathbf{x}$ and $\tilde{\mathbf{x}}$ are target and source images, $y_{\tt{tgt}}$ and $y_{\tt{src}}$ are target and source prompts.
However, we found that \ssname with InstructPix2Pix is more amenable in editing real images as it does not require source prompt. 
Finally, we remark that \ssname can be applied to various text-to-image diffusion models such as ControlNet~\citep{zhang2023adding}, which we leave it for the future work.

\section{Additional experiments}\label{appen:addexp}
\subsection{Compositional editing}
Recent works have shown the ability of text-to-image diffusion models in compositional \emph{generation} of images handling multiple prompts~\citep{du2023reduce, liu2022compositional}.
Here, we show that \ssname can extend this ability to compositional \emph{editing}, even at panorama-scale images which require a particular ability to maintain far-range consistency. Specifically, we demonstrate that one can edit a panorama image to follow different prompts on different regions while keeping the overall context uncorrupted.

Given multiple textual prompts $\{y_k\}_{k=1}^K$, the compositional noise estimate is given by
\begin{align*}
    \boldsymbol{\epsilon}_\phi(\mathbf{x}_t;\{y_k\}_{k=1}^K,t) = \sum_{k=1}^K 
    \alpha_k\boldsymbol{\epsilon}_\phi^{\omega}(\mathbf{x}_t;y_k,t),
\end{align*}
where $\alpha_k$ are hyperparameters that regularize the effect of each prompt. 
When applying compositional generation to the panorama image editing, the challenge lies in obtaining image that is smooth and natural within the region where the different prompts are applied. 
To that end, for each patch of an image, we set $\alpha_k$ to be the area of the overlapping region between the patch and region where prompt $y_k$ is applied. 
Also, we normalize to assure $\sum_{k}\alpha_k=1$. In Figure~\ref{fig:appen_comp_gen}, we illustrate some examples on compositional editing of a panorama image. For instance, given an image, one can change into different weathers, different seasons, or different painting styles without leaving artifacts that hinder the spatial consistency of an image. 

\subsection{Text-to-3D generation with \sname}\label{appen:add-dreamfusion}
As of Section~\ref{sec:ablation}, we present a detailed study on the effect of \sname 
{in text-to-3D generation, particularly focusing on the DreamFusion architecture~\citep{poole2022dreamfusion}.
We follow the most of experimental setups from those conducted by \citet{poole2022dreamfusion}. 
Our experiments in this section are based on Stable-DreamFusion \citep{stable-dreamfusion}, a public re-implementation of DreamFusion, given that currently the official implementation of DreamFusion is not available on public.}

\paragraph{Setup.} 
We use vanilla MLP based NeRF architecture~\citep{mildenhall20nerf} with 5 ResNet~\citep{he2016deep} blocks. 
Other regularizers such as shading, camera and light sampling are set as default in \citep{stable-dreamfusion}.
We use view-dependent prompting given the sampled azimuth angle and interpolate by the text embeddings. 
We use Adan~\citep{xie2022adan} optimizer with learning rate warmup over 2000 steps from $10^{-9}$ to $2\times10^{-3}$ followed by cosine decay down to $10^{-6}$. 
We use batch size of 4 and optimize for 10000 steps in total, where most of the case sufficiently converged at 7000 to 8000 steps.
For the base text-to-image diffusion model, we adopt {\tt DeepFloyd-IF-XL-v1.0} since 
{we found it way better than the default choice of Stable Diffusion in a qualitative manner.} 
While the original DreamFusion~\citep{poole2022dreamfusion} used guidance scale of 100 for their experiments, we find that guidance scale of 20 works well for {\tt DeepFloyd}.
We selected 30 prompts used in DreamFusion gallery\footnote{\url{https://dreamfusion3d.github.io/gallery.html}} and 
{compare their generation results via DreamFusion from the standard SDS and those from our proposed \sname.}
We use one A100 (80GB) GPU for each experiment, and it takes $\sim$5 hours to conduct one experiment.

For \sname implementation, we use LPIPS~\citep{zhang2018perceptual} as a distance of RBF kernel. Note that LPIPS gives more computational cost than the usual $\ell_2$-norm based RBF kernel. The LPIPS is computed between two rendered views of size 64$\times$64. For the kernel bandwidth, we use $h=\frac{ {\tt{med}}^2}{\log B}$, where $\tt{med}$ is a median of the pairwise LPIPS distance between the views, $B$ is the batch size. 

\begin{table*}[t]
\centering\small
\vspace{-0.1in}
\begin{minipage}{.4\textwidth}
\centering\small
\includegraphics[width=0.9\textwidth]{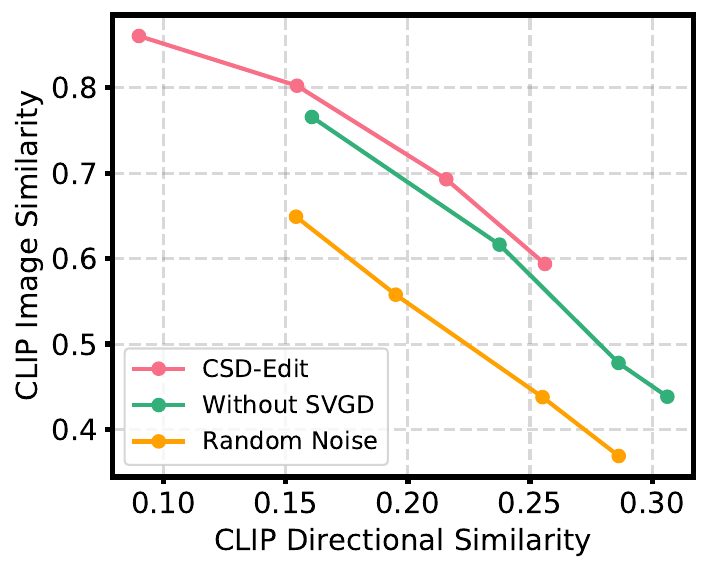}
\vspace{-0.1in}
\captionof{figure}{\textbf{Ablation study}.
Ablation study on the components of \ssname  at different guidance scales $\omega_y\in\{3.0, 5.0, 7.5, 10.0\}$.
}\label{fig:exp_ablation_pan}
\end{minipage}
\hfill
\begin{minipage}{.56\textwidth}
{
\centering\Large

\caption{\textbf{Text-to-3D}. Quantitative comparison between \sname and SDS under on text-to-3D generation via DreamFusion~\citep{poole2022dreamfusion} 
}\label{tab:appen_dreamfusion}
\resizebox{\textwidth}{!}{
        \begin{tabular}{c|c|c|c}
            \toprule 
             & {CLIP Similarity} & {CLIP Similarity} & {FID}\\ 
             & {Color} $\uparrow$ & {Geo} $\uparrow$ & $\downarrow$\\ 
             \midrule
             SDS~\citep{poole2022dreamfusion} & 0.437 & 0.322 & 259.4\\
             \rowcolor{Gray}\textbf{\sname (Ours)} & \textbf{0.447} & \textbf{0.345} & \textbf{247.1}\\
             \bottomrule
        \end{tabular}
}
}
\end{minipage}
\end{table*}

For evaluation, we render the scene at the elevation at 30 degree and capture at every 30 degree of azimuth angle. Then we compute the CLIP image-text similarity between the rendered views and input prompts. We measure similarities for both textured views (RGB) and textureless depth views (Depth). We also report Frechet Inception Distance~(FID) between the RGB images and ImageNet validation dataset to evaluate the quality and diversity of rendered images compared to natural images.
\paragraph{Results.}

In Table~\ref{tab:appen_dreamfusion}, we report the evaluation results of \sname on text-to-3D generation comparison to DreamFusion. Remark that \sname presents better CLIP image-text similarities in both RGB and Depth views. Also, \sname achieves lower FID score showing its better quality on generated samples. Since we used the same random seed in generating both \sname and DreamFusion, the shapes and colors are similar. However, the results show that \sname obtains finer details in its generations. 

In Figure~\ref{fig:appen_gen_3d}, we qualitatively compare the baseline DreamFusion (SDS) and ours. We empirically observe three benefits of using \sname over SDS. 
First, \sname provides better quality compared to SDS. SDS often suffers from Janus problem, where multiple faces appear in a 3D object. We found that \sname often resolves Janus problem by showing consistent information during training. See the first row of Figure~\ref{fig:appen_gen_3d}.
Second, \sname can give us better fine-detailed quality. The inconsistent score distillation often gives us blurry artifact or undesirable features left in the 3D object. \sname can handle this problem and results in higher-quality generation, e.g., Figure~\ref{fig:appen_gen_3d} second row.
Lastly, \sname can be used for improving diversity. One problem of DreamFusion, as acclaimed by the authors, is that it lacks sample diversity. Thus, it often relies on changing random seeds, but it largely alters the output. On the other hand, we show that \sname can obtain alternative sample with only small details changed, e.g., Figure~\ref{fig:appen_gen_3d} third row. Even when SDS is successful, \sname can be used in generating diverse sample.

\section{Ablation study}\label{appen:ablation}

In addition to the qualitative examples shown in Section~\ref{sec:ablation}, we present an additional ablation study on (a) the effect of SVGD and (b) subtracting random noise in \ssname in panorama image editing experiments.
Following the experimental setup in Section~\ref{sec:exp-panorama}, we select 16 images and apply 5 different artistic stylization using \ssname, \ssname without SVGD, and \ssname without subtracting image-conditional noise estimate. 
Again, we measure the CLIP image similarity and CLIP directional similarity for the evaluation. 

In Figure~\ref{fig:exp_ablation_pan}, we plot the results of the ablation study. Remark that \ssname without SVGD radically changes the image due to the absence of consistency regularization. As illustrated in Figure~\ref{fig:abl}, \ssname via subtracting random noise instead of image-conditional noise results in blurry outputs. Here, we also quantitatively show that it results in significant degrade in CLIP image similarity and CLIP directional similarity, losing the details of the source image. In Figure~\ref{fig:appen_abl}, we depict the qualitative results on our ablation study.

\section{Implementation details}\label{appen:impl}
\paragraph{Setup.}
For the experiments with \ssname, we use the publicly available pre-trained model of Instruct-Pix2Pix~\citep{brooks2022instructpix2pix}\footnote{\url{https://github.com/timothybrooks/instruct-pix2pix}} by default. We perform \ssname optimization on the output space of Stable Diffusion~\citep{rombach2022high} autoencoder. 
% We simply used SGD optimizer with decaying learning rate decay and did not use weight decay. 
{We use SGD optimizer with step learning rate decay, without adding weight decay.}
We set $t_{\tt{min}}=0.2$ and $t_{\tt{max}}=0.5$, where original SDS optimization for DreamFusion used $t_{\tt{min}}=0.2$ and $t_{\tt{max}}=0.98$. 
This is because we do not generally require a large scale of noise in editing. 
We use the guidance scale $\omega_y\in[3.0, 15.0]$ and image guidance scale $\omega_s\in[1.5,5.0]$. We find that our approach is less sensitive to the choice of image guidance scale, yet a smaller image guidance scale is more sensitive to editing. All experiments are conducted on AMD EPYC 7V13 64-Core Processor and a single NVIDIA A100 80GB.
Throughout the experiments, we use OpenCLIP~\citep{ilharco_gabriel_2021_5143773} {\tt{ViT-bigG-14}} model for evaluation.

\subsection{Panorama image editing}\label{appen:high_res}
To edit a panorama image, we first encode into the Stable Diffusion latent space (i.e., downscale by 8), then use a stride size of 16 to obtain multiple patches. Then we select a $B$ batch of patches to perform \ssname. Note that we perform \ssname and then normalize by the number of appearances as mentioned in Section~\ref{sec:appl}. 
Note that our approach performs well even without using small batch size, e.g., for an image of resolution 1920$\times$512, there are 12 patches and we use $B=4$. 

For experiments, we collect 32 panorama images and conduct 5 artistic stylizations: ``turn into Van Gogh style painting'', ``turn into Pablo Picasso style painting'', ``turn into Andy Warhol style painting'', ``turn into oriental style painting'', and ``turn into Salvador Dali style painting''. We use learning rate of 2.0 and image guidance scale of 1.5, and vary the guidance scale from 3.0 to 10.0. 

\subsection{Video editing}\label{appen:video}
We edit video sequences in DAVIS 2017~\citep{Pont-Tuset_arXiv_2017} by sampling 24 frames at the resolution of 1920$\times$1080 from each sequence. Then, we resize all frames into 512$\times$512 resolution and encode all frames each using Stable Diffusion. We use learning rate $[0.25, 2]$ and optimize them for $[200, 500]$ iterations.

\subsection{3D scene editing}\label{appen:3d}
Following Instruct-NeRF2NeRF~\citep{instructnerf2023}, we first pretrain NeRF using the \emph{nerfacto} model from NeRFStudio~\citep{tancik2023nerfstudio}, training it for 30,000 steps. 
Next, we re-initialize the optimizer and finetune the pre-trained NeRF model with edited train views. 
In contrast to Instruct-NeRF2NeRF, which edits one train view with Instruct-Pix2Pix after every 10 steps of update, we edit a batch of train views (batch size of 16) with \ssname after every 2000 steps of update.
The batch is randomly selected among the train views without replacement.

\newpage
\section{Additional qualitative results}\label{appen:vis}

\begin{table*}[ht]
\centering\small
\vspace{-0.07in}
\centering\small
\includegraphics[width=\textwidth]{{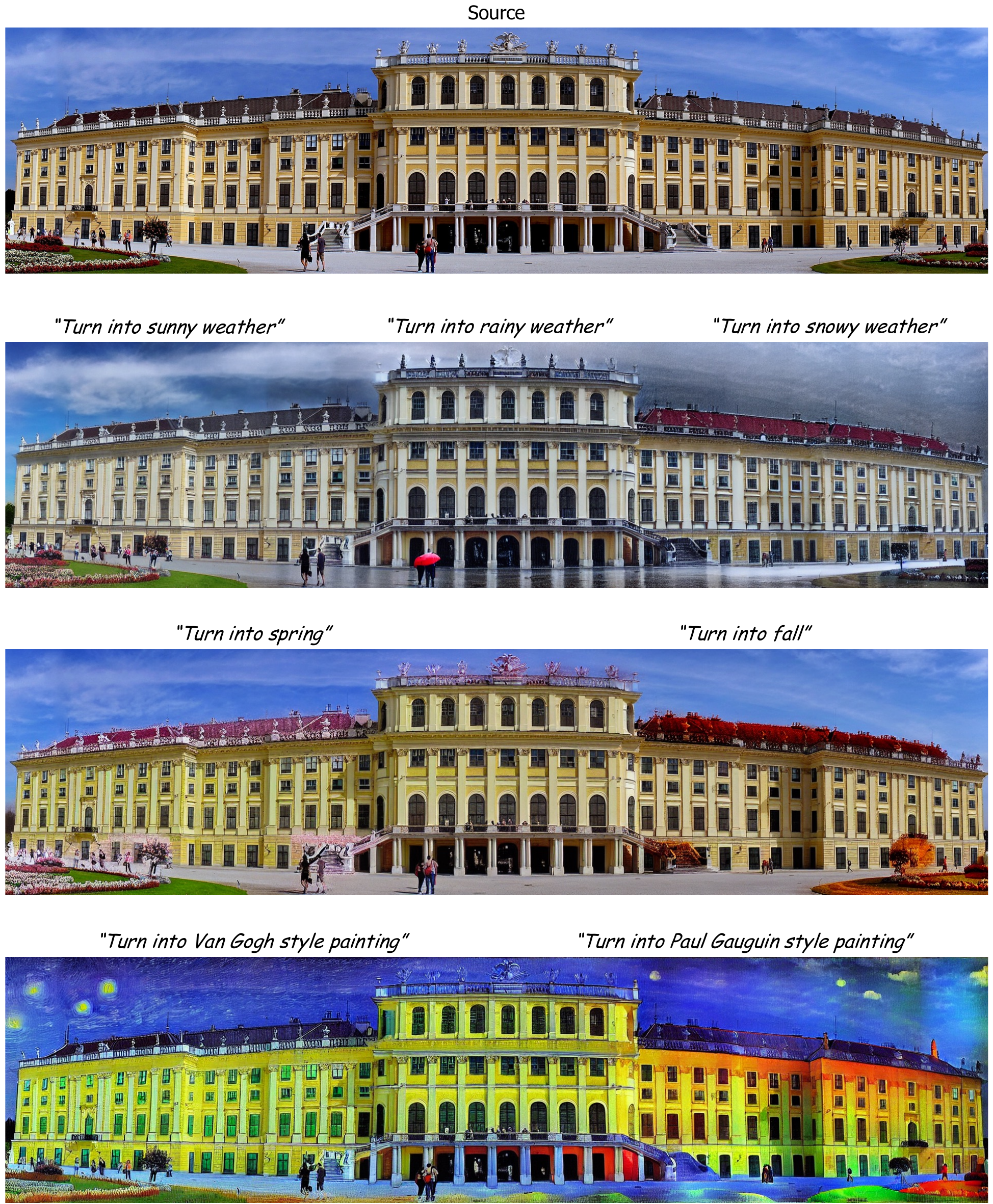}}
\captionof{figure}{\textbf{Compositional image editing}. \ssname demonstrates the ability to edit consistently and coherently across patches in panorama images. This provides the unique capability to manipulate each patch according to different instructions while maintaining the overall structure of the source image. Remarkably, \ssname ensures a smooth transition between patches, even when different instructions are applied. 
}\label{fig:appen_comp_gen}
\vspace{-10pt}
\end{table*}

\newpage
\begin{table*}[ht]
\centering\small
\vspace{-0.07in}
\centering\small
\includegraphics[width=\textwidth]{{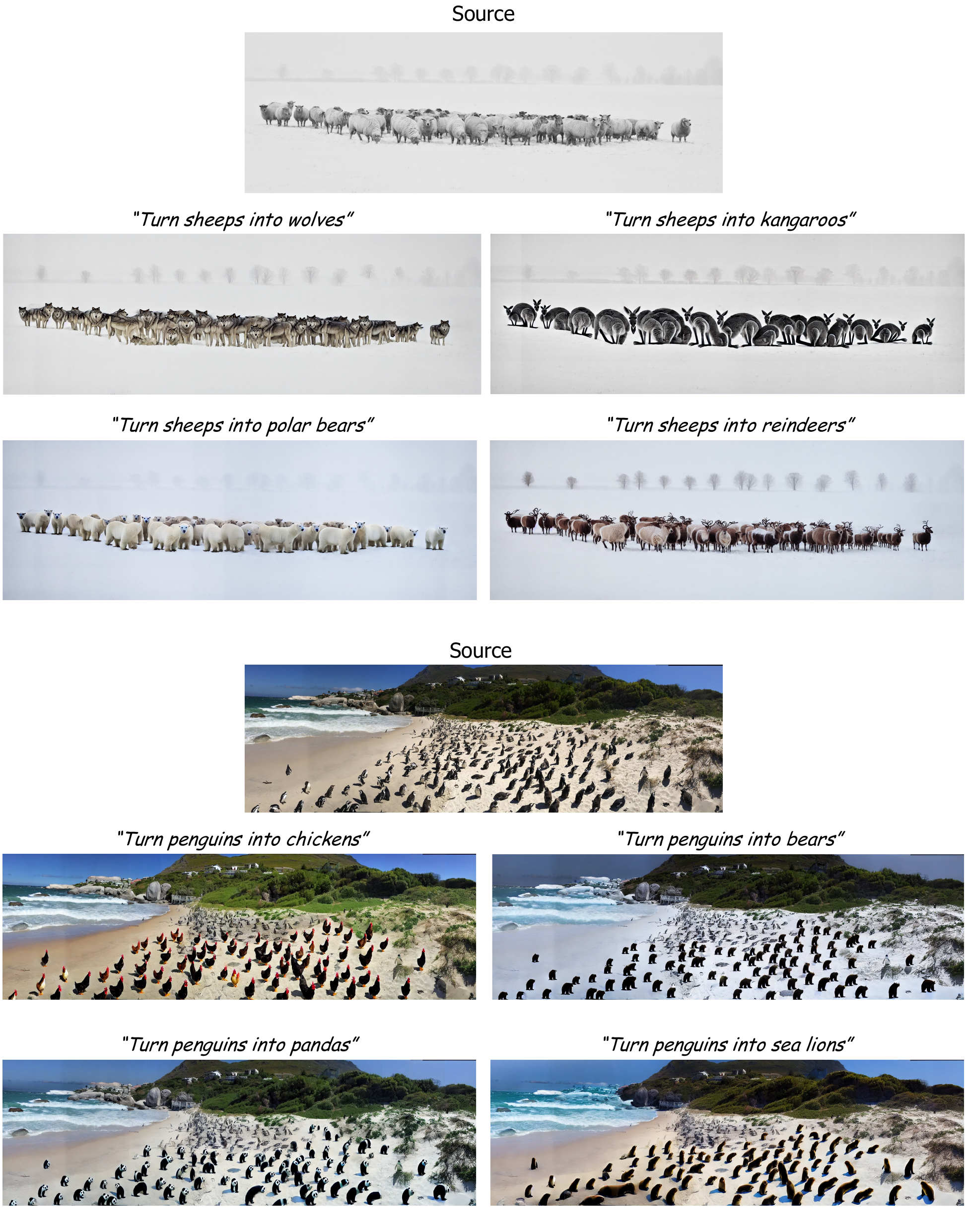}}
\captionof{figure}{\textbf{Object editing}. \ssname can edit many objects in a wide panorama image consistently in accordance with the given instruction while preserving the overall structure of source images. 
}\label{fig:appen_object}
\vspace{-10pt}
\end{table*}

\newpage
\begin{table*}[ht]
\centering\small
\vspace{-0.07in}
\centering\small
\includegraphics[width=\textwidth]{{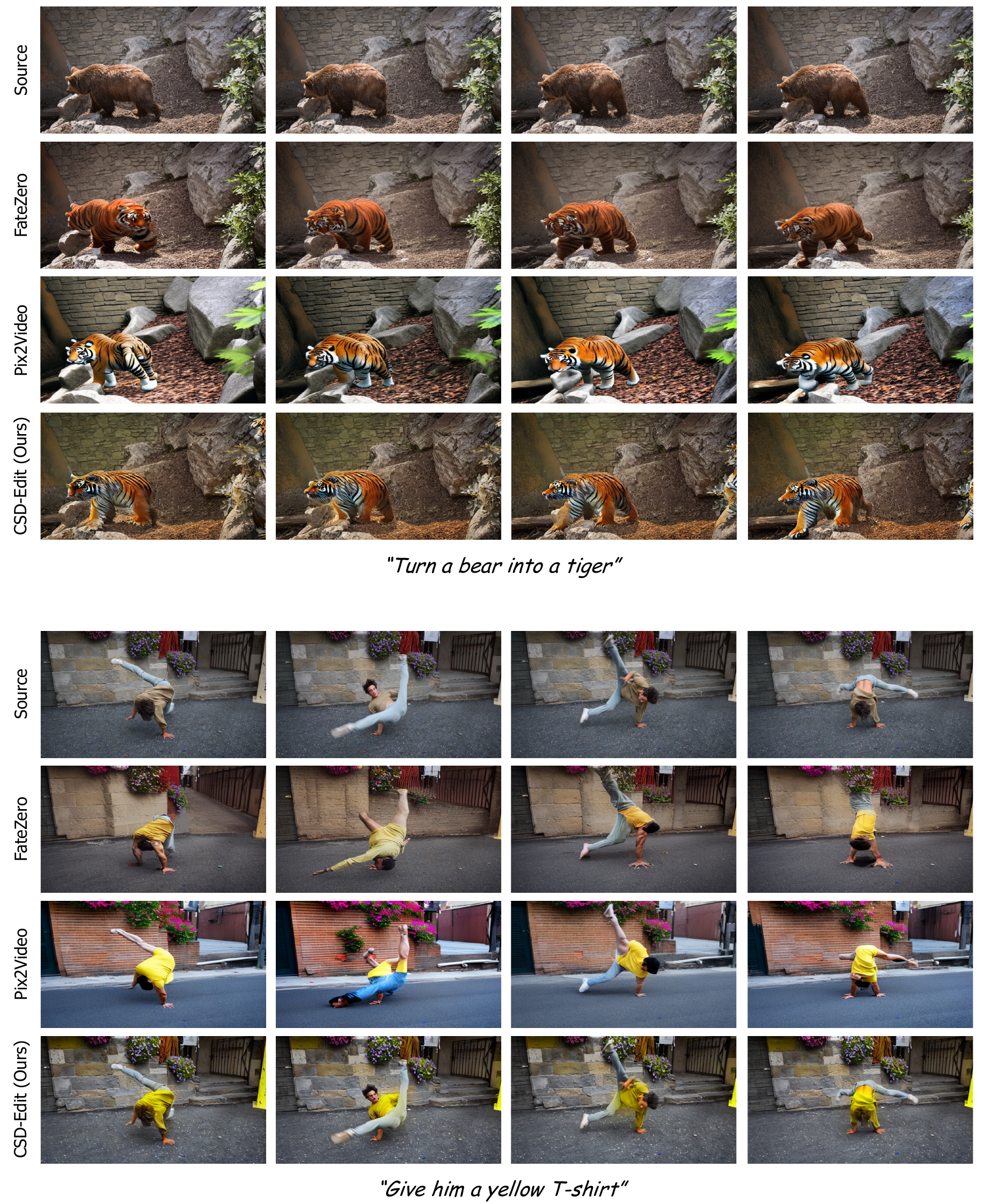}}
\captionof{figure}{\textbf{Video editing}. \ssname demonstrates various editing from an object (e.g., tiger) to attributes (e.g., color) while providing consistent edits across frames and maintaining the overall structure of a source video.
}\label{fig:appen_video}
\vspace{-10pt}
\end{table*}

\newpage
\begin{table*}[ht]
\centering\small
\vspace{-0.07in}
\centering\small
\includegraphics[width=\textwidth]{{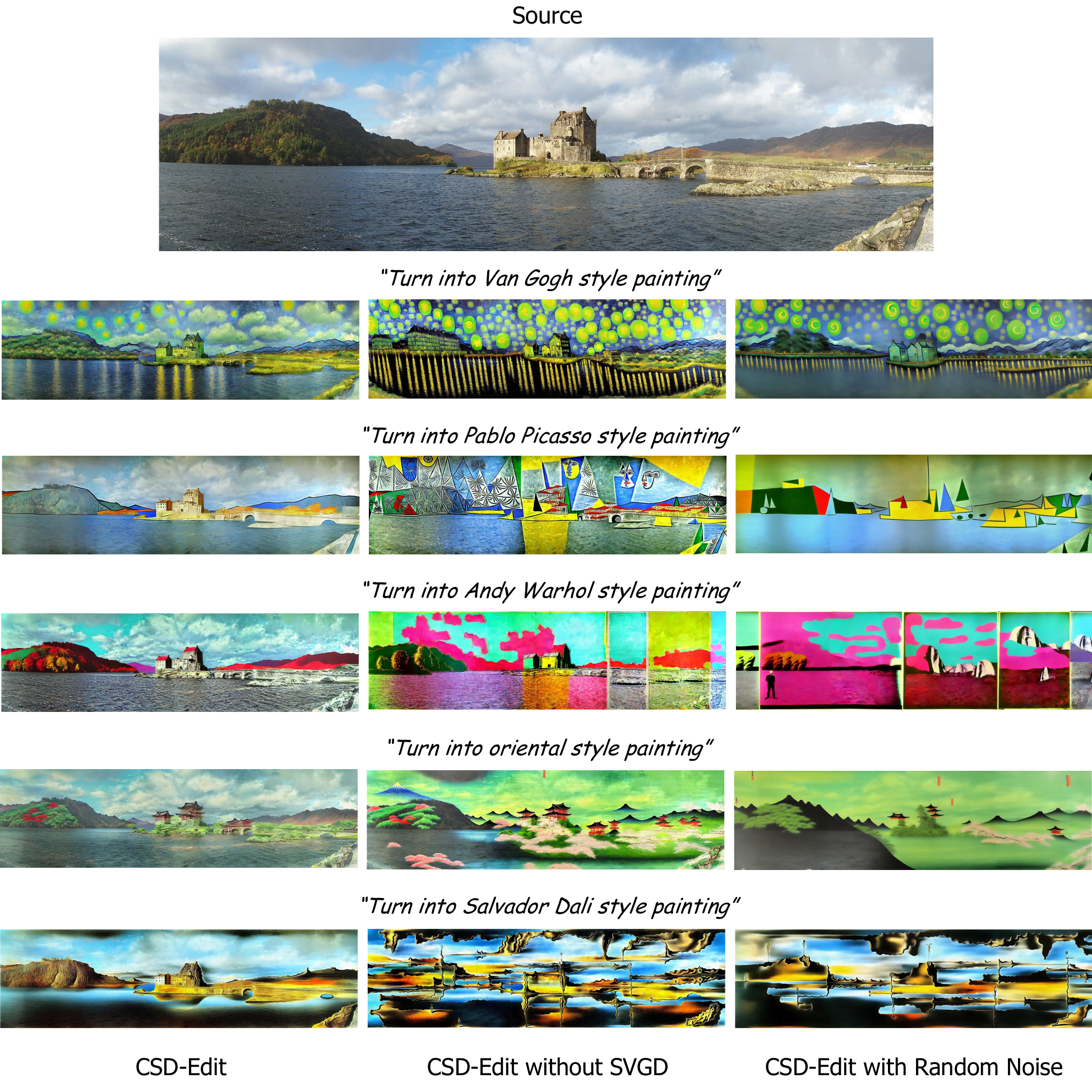}}
\captionof{figure}{
\textbf{Ablation study: SVGD and random noise}. 
% Ablation study of Stein Variational Gradient Descent (SVGD) and random noise. 
As illustrated, edits across different patches are not consistent without SVGD. Also, when using random noise as baseline noise, it loses the content and the detail of the source image. 
}\label{fig:appen_abl}
\vspace{-10pt}
\end{table*}

\newpage
\begin{table*}[ht]
\centering\small
\vspace{-0.07in}
\centering\small
\includegraphics[width=\textwidth]{{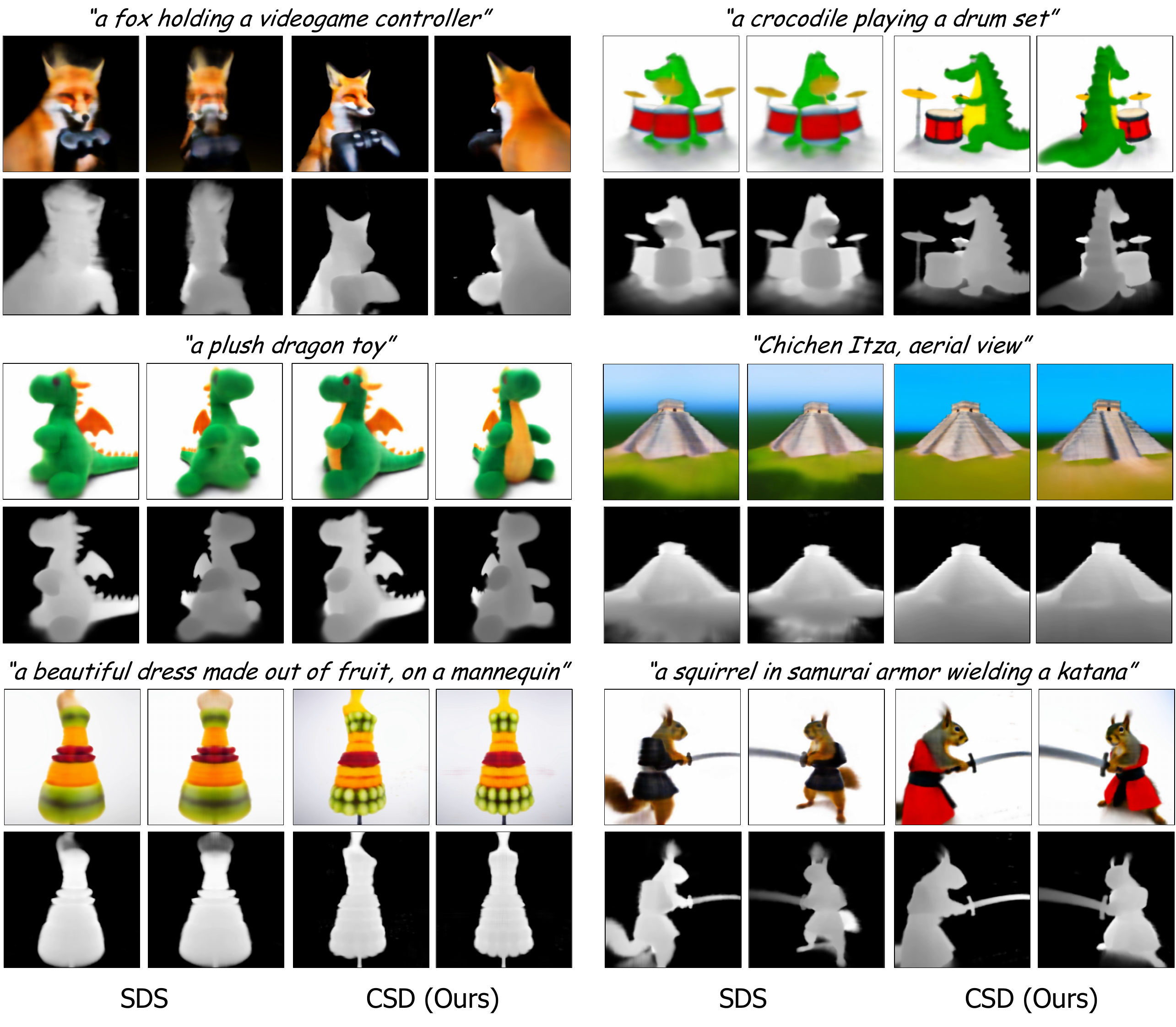}}
\captionof{figure}{\textbf{Text-to-3D generation examples}. (First row) \sname helps to capture coherent geometry compared to using SDS. (Second row) \sname allows learning finer details than SDS. (Third row) \sname can provide diverse and high-quality samples without changing random seeds.
}\label{fig:appen_gen_3d}
\vspace{-10pt}
\end{table*}

\newpage
\section{Limitations}\label{appen:limitation}
\begin{table*}[t]
\centering\small
\vspace{-0.07in}
\centering\small
\includegraphics[width=\textwidth]{{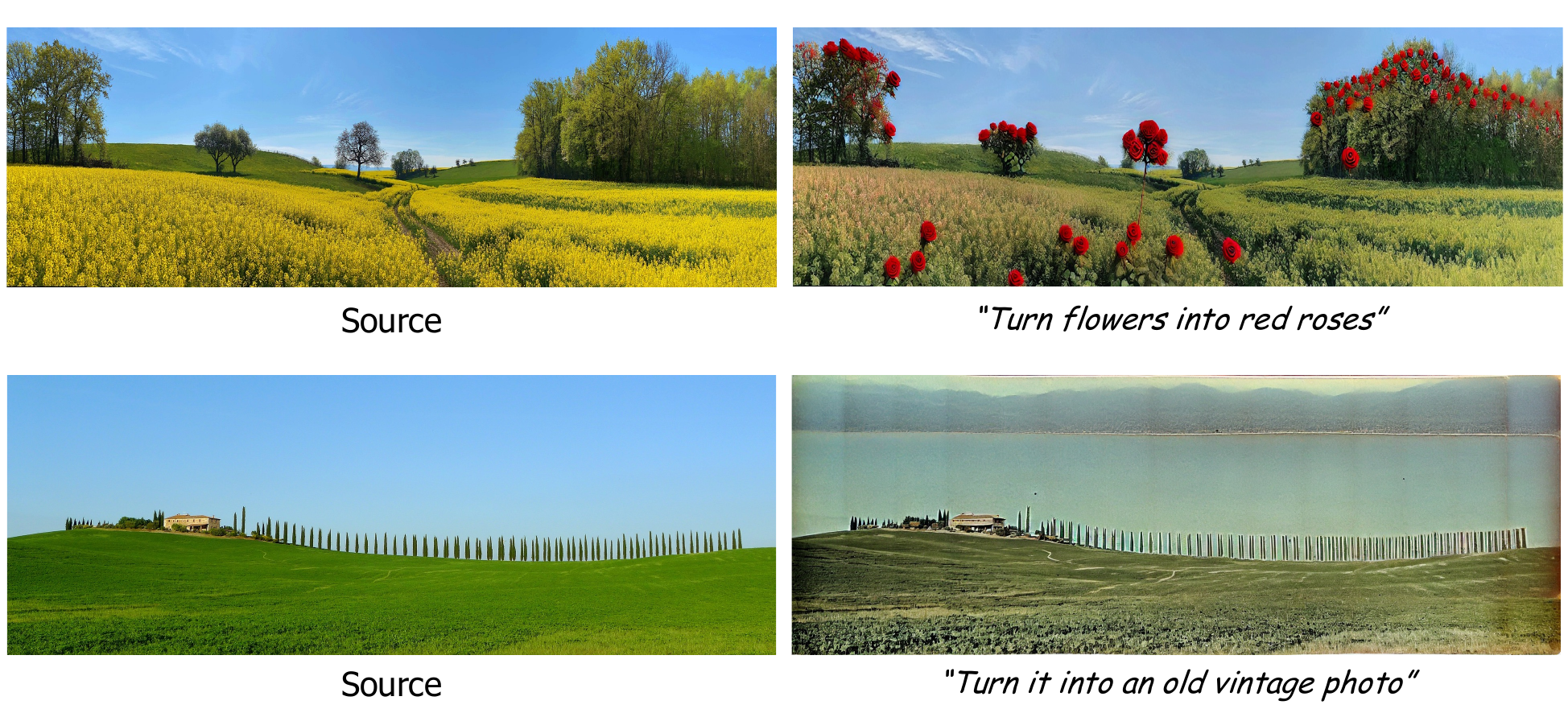}}
\captionof{figure}{\textbf{Limitations}. (First row) \ssname often manipulates undesirable contents due to the inherent inability of Instruct-Pix2Pix model. 
(Second row) \ssname often produces artifacts on the image due to the patch-wise update. 
}\label{fig:appen_limit}
\vspace{-10pt}
\end{table*}
As our method leverages pre-trained Instruct-Pix2Pix, it inherits the limitations of it such as undesirable changes to the image due to the biases. 
Also, as described in \citep{brooks2022instructpix2pix}, Instruct-Pix2Pix is often unable to change viewpoints, isolate a specific object, or reorganize objects within the image. 

When editing a high-resolution image by dividing it into patches, it often remains an artifact at the edge of the patches, especially at the corner side of an image. This is due to that the patches at the corner are less likely to be sampled during the optimization. 
See Figure~\ref{fig:appen_limit} for examples. 

When editing a video, the edited video often shows a flickering effect due to the inability of the Stable Diffusion autoencoder in compressing the video. We believe that using \ssname with video diffusion models trained on video datasets can possibly overcome this problem. 

\section{Broader Impact}\label{appen:broad}
Our research introduces a comprehensive image editing framework that encompasses various modalities, including high-resolution images, videos, and 3D scenes. While it is important to acknowledge that our framework might be potentially misused to create fake content, this concern is inherent to image editing techniques as a whole.
Furthermore, our method relies on generative priors derived from large text-to-image diffusion models, which may inadvertently contain biases due to the auto-filtering process applied to the vast training dataset. These biases influence the score distillation process, where the undesired results may come out.
However, we propose that employing Consistent Score Distillation (CSD) can assist us in identifying and understanding such undesirable biases. By leveraging the inter-sample relationships and aiming for consistent generation and manipulation of visual content, our method provides a valuable avenue for comprehending the interaction between samples and prompts. Exploring this aspect further represents an intriguing future direction. 
% \placeholder{1}
\end{document}